\pdfoutput=1
\documentclass[12pt]{article}
\usepackage[utf8]{inputenc}
\usepackage[T1]{fontenc}
\usepackage{lmodern}
\usepackage[margin=1in]{geometry}
\usepackage{setspace,authblk,fancyhdr,microtype,ragged2e}
\usepackage{amsmath,amssymb,bm,mathtools}
\usepackage{booktabs,longtable,tabularx,array,multirow,makecell,threeparttable,adjustbox}
\usepackage{graphicx,caption,subcaption,float,placeins,pdflscape,rotating}
\usepackage{algorithm,algpseudocode}
\usepackage[authoryear,round]{natbib}
\setcitestyle{citesep={,}}
\usepackage[colorlinks=true,allcolors=blue]{hyperref}
\usepackage{enumitem}

\newcommand{\method}{HJB residual}

\newcolumntype{Y}{>{\RaggedRight\arraybackslash}X}
\newcolumntype{P}[1]{>{\RaggedRight\arraybackslash}p{#1}}
\numberwithin{equation}{section}
\setlist[itemize]{leftmargin=1.3em}
\setlist[enumerate]{leftmargin=1.5em}

\title{\textbf{Autopilot-Preserving Residual Q-Learning with HJB-Inspired Finite-Action Risk Filtering for Fixed-Wing UAV Command Supervision}}
\author[1]{Mehmet ISCAN\thanks{Corresponding author: \texttt{miscan@yildiz.edu.tr}}}
\author[2]{Batuhan Temiz}
\affil[1]{PythaLab, Yildiz Technical University, Istanbul, Turkey}
\affil[2]{Autopilot Systems Chief Engineer, Turkish Aerospace (TUSA\c{S}), Ankara, Turkey, \texttt{batuhan.temiz@tai.com.tr}}
\date{}

\begin{document}
\maketitle
\noindent\textbf{Running title:} HJB-inspired residual UAV command supervision\\
\textbf{Corresponding author:} Mehmet ISCAN, PythaLab, Yildiz Technical University, Istanbul, Turkey. Email: \texttt{miscan@yildiz.edu.tr}

\begin{abstract}
A fixed-wing unmanned aerial vehicle (UAV) must hold airspeed, altitude, and heading references under wind, gusts, and turbulence, where these channels are coupled so that a correction improving one can degrade another, and classical autopilots stabilize the airframe well yet adapt poorly when a hard crosswind coincides with an aggressive turn, while reinforcement-learning (RL) policies that act directly on the flight surfaces concentrate exploration risk at the actuator interface. We place a learned supervisor above an unchanged autopilot rather than inside it: the supervisor selects a residual from a finite, bounded action set on the commanded airspeed, altitude, and heading, the modified reference is projected into an admissible command envelope before it reaches the autopilot, and the autopilot remains the only actuator-facing controller. What is new is how the residual is chosen: \method\ scores candidate residuals with a semi-discrete value-iteration critic in the spirit of the Hamilton-Jacobi-Bellman (HJB) equation, ranks them by a no-op-relative Hamiltonian advantage, and passes them through a control-Lyapunov- and control-barrier-inspired finite-action shield that always retains a no-op fallback. On a shared 12-state simulation runtime that holds the plant, autopilot, and actuator model fixed, so the comparison is at the level of the whole command-supervision package, \method\ lowers mean root-mean-square path-tracking error to 44.809 m, against 338.617 m for the baseline autopilot and 88.809 m for a tabular-Q residual supervisor, an 86.77\% reduction relative to the baseline and a 49.54\% reduction relative to Q-learning; evaluated per scenario, this gain is concentrated where the baseline fails worst and comes with a measured rise in airspeed error, so no method dominates every metric. We formulate reference tracking as autopilot-preserving residual command supervision, design the \method\ critic-and-filter supervisor that selects residuals only on airspeed, altitude, and heading, and evaluate it against a classical autopilot and a Q-only supervisor in one fixed-wing runtime with its trade-offs reported intact.
\end{abstract}


\section{Introduction}
A fixed-wing UAV cannot decouple path tracking from speed, load factor, and energy state. A heading or altitude correction that improves spatial tracking also changes airspeed regulation, throttle usage, and saturation exposure, and this coupling tightens for small and miniature aircraft. The Beard--McLain modeling basis used here separates air-relative variables, inertial motion, and ground-track behavior over a 12-state rigid body with nested longitudinal and lateral loops \citep{beard_mclain_ngc}, and wind studies show that the disturbance reshapes which paths a fixed-wing platform can fly at all: steady wind changes which maneuvers and timings are feasible \citep{ayhan2018}, while low-altitude wind shear degrades curved-path following enough to motivate dedicated guidance or estimation structure \citep{zhang2024}. Holding airspeed, altitude, and heading references under wind, gusts, and turbulence is therefore a coupled control problem, not a set of independent loops.

Classical autopilots address this problem with interpretable, stability-oriented, actuator-facing feedback. The small-UAV architecture of \citet{beard_mclain_ngc} separates guidance, path following, autopilot, airframe, and state estimation, and its successive-loop-closure design makes saturation limits and bandwidth separation explicit. Contemporary fixed-wing work keeps fixed or gain-scheduled proportional-integral-derivative structures so that controllers stay interpretable and re-tunable across airspeed regimes and remain compatible with the actuators \citep{poksawat2017,liu2015}. Such controllers give a credible plant-facing baseline, but their adaptation is scheduled on operating point rather than on the disturbance and reference combination at hand, so they leave little room to react when a hard crosswind coincides with an aggressive turn. The baseline autopilot in this study is such an engineering reference, with authority worth preserving rather than a degraded or intentionally weak baseline.

When the conditions turn hard, both purely nominal and purely learned control reach their limits, and this is where the serious problems lie. Studies that push tracking under turbulence tend to act on the inner loop, through segmented control surfaces and retuning \citep{sattar2022} or a redesigned robust nonlinear law \citep{meharie2024}; they predict the trade-offs among reference error, airspeed, energy, and load that we later observe, but they reach them by altering the low-level controller. RL policies, meanwhile, are brittle under dynamics variation and parameter uncertainty unless backed by adaptive or robust control structure \citep{annaswamy2023,cheng2022}, and the safe-RL literature identifies exploratory actions as a direct source of constraint-violation risk in real engineering systems \citep{li2018}. A controller scheduled on operating point cannot adapt to the worst disturbance-and-reference mixtures, and an actuator-level learner that could adapt puts its exploration exactly where a fixed-wing aircraft can least afford it. These observations point to an intermediate architecture: add adaptation above an existing autopilot, but keep exploratory authority away from the elevator, aileron, rudder, and throttle.

Residual and supervisory learning offer that middle path between a frozen controller and end-to-end actuator-level RL. Residual RL splits a task into a conventional feedback component and a learned correction, keeping model-based structure while learning what is hard to model exactly, although its canonical formulation injects the correction at the torque or actuator interface \citep{johannink2019}; aerospace residual learning carries this idea into flight while arguing for explicit safety machinery around the learned term \citep{jayarathne2023}, and supervisory RL in another safety-critical plant places the learned decision above the plant-facing loop rather than inside it \citep{sun2024}. A separate control-theoretic strand treats the reference and command signals themselves as the design surface, through command filtering, modified command and reference models, and command limiting \citep{dong2011,na2019,sun2021}. Together these strands justify intervening at the command layer, yet none combines a finite residual on commanded airspeed, altitude, and heading with a shared-runtime comparison against both a classical autopilot and a Q-only residual, and none reports the airspeed and control-activity cost such an intervention incurs.

Safe RL and value-guided control supply the vocabulary for filtering a finite set of residual actions while drawing their own claim boundaries. Constraint-admissible safe RL restricts exploration to a supervised action set \citep{li2018}, control Lyapunov and control barrier functions formalize stabilization and safety as pointwise conditions enforced through optimization-based filters \citep{ames2016,cohen2020,gurriet2020,zhao2023}, Hamilton-Jacobi safety analysis gives rigorous value-based safety concepts that scale poorly with state dimension \citep{fisac2019}, and Hamiltonian-driven approximate dynamic programming motivates scoring candidate actions by a value function and a Hamiltonian residual under assumptions stronger than those available here \citep{yang2017,yang2021,yang2022,lutter2020}, with learned safe-value functions extending this to critics that act as constraints under verification assumptions stronger than a coarse simulation runtime supplies \citep{tan2024}. We borrow the admissible-action, no-op-fallback, value-proxy, and risk-gate ideas from this literature as design elements without inheriting its certificates, because an exact barrier or reachability solution over a 12-state aircraft is not what a finite-grid critic delivers. The space we work in lies between a controller too rigid to adapt and a solver too expensive to run at flight dimension: no cited study jointly provides a bounded, autopilot-preserving, finite-action residual command supervisor evaluated, with its costs disclosed, against both a classical autopilot and a Q-only supervisor in one runtime.

We close that space with the \method\ supervisor. Learning acts only on the commanded airspeed, altitude, and heading; every modified command is projected into an admissible envelope before it reaches the unchanged autopilot; and a semi-discrete value-iteration critic, inspired by the HJB and Bellman equations, ranks the finite residual candidates that a control-Lyapunov- and control-barrier-style filter admits, scoring them by a no-op-relative Hamiltonian advantage and always retaining a no-op fallback. External literature motivates and contrasts the architecture, while the numerical claims rest on the simulation reported here: across a shared 12-state runtime, residual supervision lowers spatial and altitude tracking error while raising airspeed error, the \method\ critic recovers the scenarios where a Q-only residual diverges, and no method wins on every metric.

This paper makes the following contributions. We formulate fixed-wing UAV reference tracking as an autopilot-preserving residual command-supervision problem in which learning reshapes only the commanded airspeed, altitude, and heading; we design the \method\ supervisor, which scores candidate residuals with a semi-discrete HJB- and Bellman-inspired value critic, ranks them by a no-op-relative Hamiltonian advantage, and filters them through control-Lyapunov- and control-barrier-style gates that always retain a no-op fallback; and we evaluate the baseline autopilot, a tabular-Q residual supervisor, and \method\ in a shared 12-state fixed-wing runtime that holds the plant, autopilot, and actuator model common across methods, so the comparison is at the level of the controller package as a whole (the two residual supervisors also share a deterministic energy-allocation helper absent from the baseline), reporting the comparison with its trade-offs intact, so that residual supervision is shown to lower spatial and altitude tracking error while raising airspeed error, with no method dominating every metric.


\section{System model}
\label{sec:model}

\subsection{Modeling basis}
The plant is a nonlinear six-degree-of-freedom fixed-wing rigid-body model with a twelve-component state. Its state definition, coordinate frames, body--inertial transformations, force and moment decomposition, wind-relative air-data, and successive-loop-closure autopilot follow the small-UAV framework of \citet{beard_mclain_ngc}. On that basis we instantiate a coefficient-based simulation, fixing a small set of well-defined choices that the rest of this section makes explicit. Air density varies with altitude through an ISA-like ratio rather than being held at its sea-level value. The steady-wind and gust components are combined in the body frame through the textbook transformation. The motor gain and the nominal trim airspeed take stress-test values that we treat as an operating point rather than as aircraft limits. The runtime moment vector equals the aerodynamic moment, so neither propeller reaction torque nor a gyroscopic term enters the body-rate dynamics.

The state vector is
\begin{equation}
    x=
    \begin{bmatrix}
    p_n & p_e & p_d & u & v & w & \phi & \theta & \psi & p & q & r
    \end{bmatrix}^{\top},
    \label{eq:state_vector}
\end{equation}
ordered as the north, east, and down inertial coordinates \(p_n,p_e,p_d\), the body-frame translational velocities \(u,v,w\), the roll, pitch, and yaw Euler angles \(\phi,\theta,\psi\), and the body-frame angular rates \(p,q,r\). The attitude follows the 321 (yaw--pitch--roll) Euler convention, and the down coordinate fixes altitude through
\begin{equation}
    h=-p_d .
    \label{eq:altitude}
\end{equation}

The rotation \(R_b^v(\phi,\theta,\psi)\) maps a vector resolved in the vehicle frame, aligned with NED, into body-frame components, so \(a^b=R_b^v a^v\). With \(c_\phi=\cos\phi\), \(s_\phi=\sin\phi\), and analogous notation for \(\theta\) and \(\psi\),
\begin{equation}
R_b^v=
\begin{bmatrix}
 c_\theta c_\psi & c_\theta s_\psi & -s_\theta\\
 s_\phi s_\theta c_\psi-c_\phi s_\psi & s_\phi s_\theta s_\psi+c_\phi c_\psi & s_\phi c_\theta\\
 c_\phi s_\theta c_\psi+s_\phi s_\psi & c_\phi s_\theta s_\psi-s_\phi c_\psi & c_\phi c_\theta
\end{bmatrix},
\label{eq:rotation_body_vehicle}
\end{equation}
and \(R_v^b=(R_b^v)^\top\).  The translational kinematics are
\begin{equation}
\begin{bmatrix}\dot p_n\\\dot p_e\\\dot p_d\end{bmatrix}
=R_v^b(\phi,\theta,\psi)
\begin{bmatrix}u\\v\\w\end{bmatrix},
\label{eq:position_kinematics}
\end{equation}
and the Euler-angle kinematics are
\begin{equation}
\begin{bmatrix}\dot\phi\\\dot\theta\\\dot\psi\end{bmatrix}
=
\begin{bmatrix}
1 & s_\phi\tan\theta & c_\phi\tan\theta\\
0 & c_\phi & -s_\phi\\
0 & s_\phi\sec\theta & c_\phi\sec\theta
\end{bmatrix}
\begin{bmatrix}p\\q\\r\end{bmatrix}.
\label{eq:euler_kinematics}
\end{equation}
Equations~\eqref{eq:rotation_body_vehicle}--\eqref{eq:euler_kinematics} are nonlinear through the trigonometric attitude coupling, and the 321 parameterization carries the standard singularity at \(|\theta|=\pi/2\). The benchmark trajectories stay away from this pitch attitude.

\subsection{Rigid-body dynamics}
The translational and rotational dynamics close the kinematics by relating accelerations to the applied wrench. Let \(m_a\) denote mass, \(\mathbf F_b=[F_x,F_y,F_z]^\top\) the total applied force expressed in the body frame, and \(\boldsymbol\omega_b=[p,q,r]^\top\).  Newton's translational equation in the rotating body frame gives
\begin{equation}
\begin{bmatrix}\dot u\\\dot v\\\dot w\end{bmatrix}
=
\begin{bmatrix}
rv-qw\\
pw-ru\\
qu-pv
\end{bmatrix}
+\frac{1}{m_a}
\begin{bmatrix}F_x\\F_y\\F_z\end{bmatrix}.
\label{eq:translational_dynamics}
\end{equation}
The first vector on the right side is the rotating-frame coupling \(-\boldsymbol\omega_b\times[u,v,w]^\top\), and the second is the acceleration produced by the applied body forces.

Let \(\mathbf M_b=[\ell,m_y,n]^\top\) be the externally applied roll, pitch, and yaw moment vector.  For a vehicle approximately symmetric about the body \(x_b-z_b\) plane,
\begin{equation}
J=
\begin{bmatrix}
J_x & 0 & -J_{xz}\\
0 & J_y & 0\\
-J_{xz} & 0 & J_z
\end{bmatrix},
\qquad
\Gamma=J_xJ_z-J_{xz}^{2}.
\label{eq:inertia_matrix}
\end{equation}
The body-rate dynamics are
\begin{equation}
\begin{bmatrix}\dot p\\\dot q\\\dot r\end{bmatrix}
=
\begin{bmatrix}
\Gamma_1pq-\Gamma_2qr+\Gamma_3\ell+\Gamma_4n\\
\Gamma_5pr-\Gamma_6(p^2-r^2)+m_y/J_y\\
\Gamma_7pq-\Gamma_1qr+\Gamma_4\ell+\Gamma_8n
\end{bmatrix},
\label{eq:rotational_dynamics}
\end{equation}
with
\begin{align}
\Gamma_1&=\frac{J_{xz}(J_x-J_y+J_z)}{\Gamma}, &
\Gamma_2&=\frac{J_z(J_z-J_y)+J_{xz}^{2}}{\Gamma}, &
\Gamma_3&=\frac{J_z}{\Gamma}, &
\Gamma_4&=\frac{J_{xz}}{\Gamma}, \label{eq:gamma_first}\\
\Gamma_5&=\frac{J_z-J_x}{J_y}, &
\Gamma_6&=\frac{J_{xz}}{J_y}, &
\Gamma_7&=\frac{(J_x-J_y)J_x+J_{xz}^{2}}{\Gamma}, &
\Gamma_8&=\frac{J_x}{\Gamma}. \label{eq:gamma_second}
\end{align}
These equations are nonlinear in several distinct ways. Gravity enters through the attitude, the angular-rate coupling is quadratic, the aerodynamic coefficients depend on the wind-relative flow, and throttle enters thrust quadratically. The autopilot then couples the speed, altitude, and heading channels through the closed loop. This coupling is what makes a single per-channel reference adjustment propagate across the tracked outputs, and it motivates the wind-relative description that follows.

\subsection{Wind-relative air-data and atmosphere}
Aerodynamic forces act on the velocity relative to the surrounding air mass, not on the inertial velocity. The wind is decomposed into a steady part and a gust part, ordered as \([w_{n,s},w_{e,s},w_{d,s},u_{w,g},v_{w,g},w_{w,g}]\). The steady component resolves in the vehicle frame as
\begin{equation}
\mathbf V_{w,s}^{v}=\begin{bmatrix}w_{n,s}&w_{e,s}&w_{d,s}\end{bmatrix}^{\top},
\label{eq:steady_wind}
\end{equation}
and the body-frame gust component is
\begin{equation}
\mathbf V_{w,g}^{b}=\begin{bmatrix}u_{w,g}&v_{w,g}&w_{w,g}\end{bmatrix}^{\top}.
\label{eq:gust_wind}
\end{equation}
The gust is generated by a discrete-time Dryden turbulence filter driven by white noise, so the air-data signals carry the spectral content the autopilot and supervisor must reject. The total body-frame wind combines the rotated steady component with the gust,
\begin{equation}
\mathbf V_w^{b}=R_b^v(\phi,\theta,\psi)\mathbf V_{w,s}^{v}+\mathbf V_{w,g}^{b}.
\label{eq:body_wind}
\end{equation}
Subtracting this wind from the body-frame inertial velocity gives the wind-relative velocity,
\begin{equation}
\mathbf V_a^{b}=\begin{bmatrix}u_r\\v_r\\w_r\end{bmatrix}
=
\begin{bmatrix}u\\v\\w\end{bmatrix}
-
\begin{bmatrix}u_w\\v_w\\w_w\end{bmatrix}.
\label{eq:relative_air_velocity}
\end{equation}
The airspeed, angle of attack, and sideslip angle follow from \(\mathbf V_a^{b}=[u_r,v_r,w_r]^\top\) as
\begin{align}
V_a &= \sqrt{u_r^2+v_r^2+w_r^2}, \label{eq:airspeed}\\
\alpha &= \operatorname{atan2}(w_r,u_r), \label{eq:alpha}\\
\beta &= \operatorname{atan2}\!\left(v_r,\sqrt{u_r^2+w_r^2}\right). \label{eq:beta}
\end{align}
The runtime density is scaled with altitude through
\begin{equation}
\rho(h)=\max\!\left(0.05,\rho_0\,\Theta(h)^{g/(R_{air}L)-1}\right),
\quad
\Theta(h)=\frac{\max(216.65,288.15-0.0065h)}{288.15},
\label{eq:atmosphere_density}
\end{equation}
over the low-altitude tropospheric regime that the benchmark exercises, with \(\rho_0\) the sea-level reference density, \(L\) the temperature lapse rate, \(R_{air}\) the specific gas constant, and \(\Theta(h)\) the ISA temperature ratio. The density enters every aerodynamic and propulsion term through the dynamic pressure, which carries the wind-relative quantities into the forces and moments below.

\subsection{Forces and moments}
The applied wrench \([F_x,F_y,F_z,\ell,m_y,n]\) gathers gravity, aerodynamics, and propulsion. The total force and moment decompose as
\begin{align}
\mathbf F_b &= \mathbf F_g+\mathbf F_a+\mathbf F_p, \label{eq:total_force}\\
\mathbf M_b &= \mathbf M_a. \label{eq:total_moment}
\end{align}
Equation~\eqref{eq:total_moment} sets the applied moment equal to the aerodynamic moment, with no propeller reaction torque or gyroscopic contribution. Gravity in body axes is
\begin{equation}
\mathbf F_g=
\begin{bmatrix}
-m_ag\sin\theta\\
m_ag\cos\theta\sin\phi\\
m_ag\cos\theta\cos\phi
\end{bmatrix}.
\label{eq:gravity_force}
\end{equation}
Let \(\bar q=\frac{1}{2}\rho(h)V_a^2\) be the dynamic pressure, \(S\) the wing area, \(b\) the wingspan, and \(c\) the mean aerodynamic chord. The lift and drag coefficients use the Beard--McLain sigmoid stall blend, which interpolates between the linear pre-stall regime and a flat-plate post-stall model through the blending function \(\sigma(\alpha)\):
\begin{align}
\sigma(\alpha)&=\frac{1+\exp[-M(\alpha-\alpha_0)]+\exp[M(\alpha+\alpha_0)]}{(1+\exp[-M(\alpha-\alpha_0)])(1+\exp[M(\alpha+\alpha_0)])}, \label{eq:stall_sigma}\\
C_L(\alpha)&=(1-\sigma(\alpha))(C_{L0}+C_{L\alpha}\alpha)+\sigma(\alpha)\,2\operatorname{sgn}(\alpha)\sin^2\alpha\cos\alpha, \label{eq:lift_coeff}\\
C_D(\alpha)&=C_{D0}+(1-\sigma(\alpha))\epsilon(C_{L0}+C_{L\alpha}\alpha)^2+\sigma(\alpha)\,2\operatorname{sgn}(\alpha)\sin^3\alpha. \label{eq:drag_coeff}
\end{align}
Rotating lift and drag through the angle of attack gives the body-axis force coefficients
\begin{align}
C_X(\alpha)&=-C_D(\alpha)\cos\alpha+C_L(\alpha)\sin\alpha, \label{eq:cx}\\
C_Z(\alpha)&=-C_D(\alpha)\sin\alpha-C_L(\alpha)\cos\alpha. \label{eq:cz}
\end{align}
The aerodynamic force vector is
\begin{equation}
\mathbf F_a=\bar qS
\begin{bmatrix}
C_X(\alpha)+(-C_{Dq}\cos\alpha+C_{Lq}\sin\alpha)\frac{c}{2V_a}q+(-C_{D\delta_e}\cos\alpha+C_{L\delta_e}\sin\alpha)\delta_e\\
C_{Y0}+C_{Y\beta}\beta+(C_{Yp}p+C_{Yr}r)\frac{b}{2V_a}+C_{Y\delta_a}\delta_a+C_{Y\delta_r}\delta_r\\
C_Z(\alpha)+(-C_{Dq}\sin\alpha-C_{Lq}\cos\alpha)\frac{c}{2V_a}q+(-C_{D\delta_e}\sin\alpha-C_{L\delta_e}\cos\alpha)\delta_e
\end{bmatrix}.
\label{eq:aero_force}
\end{equation}
The aerodynamic moment vector is
\begin{equation}
\mathbf M_a=\bar qS
\begin{bmatrix}
b\left(C_{\ell0}+C_{\ell\beta}\beta+(C_{\ell p}p+C_{\ell r}r)\frac{b}{2V_a}+C_{\ell\delta_a}\delta_a+C_{\ell\delta_r}\delta_r\right)\\
c\left(C_{M0}+C_{M\alpha}\alpha+C_{Mq}\frac{c}{2V_a}q+C_{M\delta_e}\delta_e\right)\\
b\left(C_{n0}+C_{n\beta}\beta+(C_{np}p+C_{nr}r)\frac{b}{2V_a}+C_{n\delta_a}\delta_a+C_{n\delta_r}\delta_r\right)
\end{bmatrix}.
\label{eq:aero_moment}
\end{equation}
A momentum-disk propeller contributes thrust along the body \(x\)-axis,
\begin{equation}
\mathbf F_p=\frac{1}{2}\rho(h)S_{prop}C_{prop}
\begin{bmatrix}
(k_{motor}\delta_t)^2-V_a^2\\0\\0
\end{bmatrix}.
\label{eq:prop_force}
\end{equation}
The actuator vector is \(\delta=[\delta_e,\delta_a,\delta_r,\delta_t]^\top\), collecting elevator, aileron, rudder, and throttle. The rate-dependent terms in \(1/(2V_a)\) are set to zero when \(V_a\) falls numerically near zero, which removes the low-speed singularity. The force and moment equations use the coefficients in Tables~\ref{tab:model_parameters} and~\ref{tab:aero_coefficients}.

\begingroup
\footnotesize
\setlength{\tabcolsep}{4pt}
\renewcommand{\arraystretch}{1.08}
\begin{longtable}{P{0.13\textwidth}P{0.18\textwidth}P{0.34\textwidth}P{0.23\textwidth}}
\caption{Physical, geometric, propulsion, and state-machine parameters of the small-MAV configuration, with symbols, values, units, and origin. The model structure follows the Beard--McLain small-UAV framework.}\label{tab:model_parameters}\\
\toprule
Symbol & Value & Meaning & Origin/status\\
\midrule
\endfirsthead
\toprule
Symbol & Value & Meaning & Origin/status\\
\midrule
\endhead
$g$ & 9.8 m/s$^2$ & gravitational acceleration & project parameter\\
$m_a$ & 1.56 kg & airframe mass & Beard-McLain App. C\\
$J_x$ & 0.1147 kg m$^2$ & roll inertia & Beard-McLain App. C\\
$J_y$ & 0.0576 kg m$^2$ & pitch inertia & Beard-McLain App. C\\
$J_z$ & 0.1712 kg m$^2$ & yaw inertia & Beard-McLain App. C\\
$J_{xz}$ & 0.0015 kg m$^2$ & roll-yaw product of inertia & Beard-McLain App. C\\
$S$ & 0.2589 m$^2$ & wing planform area & Beard-McLain App. C\\
$b$ & 1.4224 m & wingspan & Beard-McLain App. C\\
$c$ & 0.3302 m & mean aerodynamic chord & Beard-McLain App. C\\
$S_{prop}$ & 0.0314 m$^2$ & propeller disk area & Beard-McLain App. C\\
$\rho_0$ & 1.2682 kg/m$^3$ & nominal sea-level density & Beard-McLain App. C, altitude-scaled at runtime\\
$M$ & 50.0 & stall-blending transition rate & Beard-McLain App. C\\
$\epsilon$ & 0.1592 & induced-drag coefficient in nonlinear drag model & Beard-McLain App. C\\
$\alpha_0$ & 0.4712 rad & stall-blending cutoff angle & Beard-McLain App. C\\
$C_{prop}$ & 1.0 & propeller force coefficient & Beard-McLain App. C\\
$k_{motor}$ & 240.0 & motor/propulsion gain used by runtime & project override for high-speed trim margin\\
$T_s$ & 0.01 s & autopilot sample time & project runtime setting\\
$\tau_f$ & 5.0 s & PID derivative filter time constant & project runtime setting\\
$V_{a,0}$ & 140.0 m/s & nominal trim/initial airspeed for benchmark runtime & project benchmark setting\\
$h_{TO}$ & 10.0 m & takeoff-zone altitude threshold & project longitudinal state machine\\
$h_{hold}$ & 10.0 m & altitude-hold deadband & project longitudinal state machine\\
$V_{rot}$ & 65.0 m/s & takeoff rotation-speed threshold & project longitudinal state machine\\
$V_{LO}$ & 78.0 m/s & liftoff-speed threshold & project longitudinal state machine\\
$h_{climbout}$ & 25.0 m & climbout transition altitude & project longitudinal state machine\\
\bottomrule
\end{longtable}
\endgroup

\begingroup
\footnotesize
\setlength{\tabcolsep}{4pt}
\renewcommand{\arraystretch}{1.08}
\begin{longtable}{P{0.15\textwidth}P{0.11\textwidth}P{0.37\textwidth}P{0.25\textwidth}}
\caption{Aerodynamic coefficients of the small-MAV configuration entering the force and moment equations, with symbols, dimensionless values, and origin. The coefficient set follows the Beard--McLain small-UAV framework.}\label{tab:aero_coefficients}\\
\toprule
Symbol & Value & Meaning & Origin/status\\
\midrule
\endfirsthead
\toprule
Symbol & Value & Meaning & Origin/status\\
\midrule
\endhead
$C_{L0}$ & 0.28 & zero-lift lift coefficient & Beard-McLain App. C\\
$C_{L\alpha}$ & 3.45 & lift slope & Beard-McLain App. C\\
$C_{Lq}$ & 0.0 & pitch-rate lift derivative & Beard-McLain App. C\\
$C_{L\delta_e}$ & -0.36 & elevator lift derivative & Beard-McLain App. C\\
$C_{D0}$ & 0.03 & zero-lift drag coefficient & Beard-McLain App. C\\
$C_{Dq}$ & 0.0 & pitch-rate drag derivative & Beard-McLain App. C\\
$C_{D\delta_e}$ & 0.0 & elevator drag derivative & Beard-McLain App. C\\
$C_{M0}$ & 0.0 & zero-angle pitching moment & Beard-McLain App. C\\
$C_{M\alpha}$ & -0.38 & pitch static stability derivative & Beard-McLain App. C\\
$C_{Mq}$ & -3.6 & pitch damping derivative & Beard-McLain App. C\\
$C_{M\delta_e}$ & -0.5 & elevator pitching-moment derivative & Beard-McLain App. C\\
$C_{Y0}$ & 0.0 & zero-sideslip side-force coefficient & Beard-McLain App. C\\
$C_{Y\beta}$ & -0.98 & side-force sideslip derivative & Beard-McLain App. C\\
$C_{Yp}$ & -0.26 & roll-rate side-force derivative & project-specific value\\
$C_{Yr}$ & 0.0 & yaw-rate side-force derivative & Beard-McLain App. C\\
$C_{Y\delta_a}$ & 0.0 & aileron side-force derivative & Beard-McLain App. C\\
$C_{Y\delta_r}$ & -0.17 & rudder side-force derivative & Beard-McLain App. C\\
$C_{\ell0}$ & 0.0 & zero-sideslip roll moment & Beard-McLain App. C\\
$C_{\ell\beta}$ & -0.12 & roll static stability derivative & Beard-McLain App. C\\
$C_{\ell p}$ & -0.26 & roll damping derivative & Beard-McLain App. C\\
$C_{\ell r}$ & 0.14 & yaw-rate roll-moment derivative & Beard-McLain App. C\\
$C_{\ell\delta_a}$ & 0.08 & aileron roll-moment derivative & Beard-McLain App. C\\
$C_{\ell\delta_r}$ & 0.105 & rudder roll-moment derivative & Beard-McLain App. C\\
$C_{n0}$ & 0.0 & zero-sideslip yaw moment & Beard-McLain App. C\\
$C_{n\beta}$ & 0.25 & yaw static stability derivative & Beard-McLain App. C\\
$C_{np}$ & 0.022 & roll-rate yaw-moment derivative & Beard-McLain App. C\\
$C_{nr}$ & -0.35 & yaw damping derivative & Beard-McLain App. C\\
$C_{n\delta_a}$ & 0.06 & aileron yaw-moment derivative & Beard-McLain App. C\\
$C_{n\delta_r}$ & -0.032 & rudder yaw-moment derivative & Beard-McLain App. C\\
\bottomrule
\end{longtable}
\endgroup


\subsection{Actuators, integration, and command interface}
The autopilot command \(\delta_{c,k}\) does not reach the airframe directly. It first passes through an actuator model that saturates the command, limits its rate, and applies a first-order lag,
\begin{align}
\delta_{c,k}^{\mathrm{clip}}&=\operatorname{clip}(\delta_{c,k},\delta_{\min},\delta_{\max}), \\
\dot\delta_k^{\mathrm{cmd}}&=\operatorname{clip}\!\left(\frac{\delta_{c,k}^{\mathrm{clip}}-\delta_k}{\tau_\delta},-\dot\delta_{\max},\dot\delta_{\max}\right), \\
\delta_{k+1}&=\operatorname{clip}(\delta_k+\Delta t\,\dot\delta_k^{\mathrm{cmd}},\delta_{\min},\delta_{\max}).
\label{eq:actuator_lag}
\end{align}
The position limits are \(\delta_e,\delta_a\in[-45^\circ,45^\circ]\), \(\delta_r\in[-30^\circ,30^\circ]\), and \(\delta_t\in[0,1]\). The corresponding rate limits are \(120^\circ/s\), \(160^\circ/s\), \(120^\circ/s\), and \(1.8/s\), and the lag time constant is \(\tau_\delta=0.08\) s. With the actuator and wind values held constant across the substeps, the twelve-state differential equation advances by fourth-order Runge--Kutta integration,
\begin{equation}
 x_{k+1}=x_k+\frac{\Delta t}{6}(k_1+2k_2+2k_3+k_4),
\label{eq:rk4}
\end{equation}
where the \(k_i\) are the standard intermediate slopes of \(\dot x=f(x,\delta,w)\) across a substep. The default step is \(T_s=0.01\) s with five Runge--Kutta substeps. Every controller mode in the benchmark shares this actuator and plant update, so the modes differ only in how they shape the reference command.

That reference command originates with the mission generator, which at decision step \(k\) provides
\begin{equation}
    r_k=\begin{bmatrix}V_{a,c,k}&h_{c,k}&\psi_{c,k}\end{bmatrix}^{\top},
    \label{eq:nominal_command}
\end{equation}
where \(V_{a,c,k}\) is the commanded airspeed, \(h_{c,k}\) the commanded altitude, and \(\psi_{c,k}\) the commanded heading. The heading command is held in degrees and converted to radians inside the autopilot, which tracks the yaw heading \(\psi\) rather than the ground-track course. A residual action \(a_k\in\mathcal A_r\) maps to the command increment
\begin{equation}
    \Delta r(a_k)=
    \begin{bmatrix}\Delta V_a(a_k)&\Delta h(a_k)&\Delta\psi(a_k)\end{bmatrix}^{\top}.
    \label{eq:residual_map}
\end{equation}
The command passed to the autopilot is
\begin{equation}
    \tilde r_k=\Pi_{\mathcal C}\left(r_k+\Delta r(a_k)\right),
    \label{eq:projected_command}
\end{equation}
with
\begin{equation}
\mathcal C=[20,140]~\mathrm{m/s}\times[0,450]~\mathrm{m}\times[-180^\circ,180^\circ).
\label{eq:command_set}
\end{equation}
The projection in Eq.~\eqref{eq:projected_command} keeps the residual-adjusted command inside \(\mathcal C\), and both residual supervisors share the same bounds. The operating point fixes the nominal trim airspeed at \(V_{a,0}=140\) m/s and the motor gain at \(k_{motor}=240\). Because the projection constrains the reference command alone, the realized trajectory still follows from the plant, the disturbances, the actuator limits, and the autopilot response.

The residual supervisor shapes this command against several competing objectives, summarized by
\begin{equation}
J=\mathbb E\!\left[\sum_{k=0}^{T}\gamma^k\left(w_r e_{\mathrm{ref},k}^2+w_h e_{h,k}^2+w_V e_{V,k}^2+w_u\|\Delta r(a_k)\|_1+w_s\sigma_k+w_n\rho_{n,k}\right)\right],
\label{eq:objective}
\end{equation}
where \(e_{\mathrm{ref},k}\) is the spatial reference error, \(e_{h,k}=h_k-h_{c,k}\) the altitude error, and \(e_{V,k}=V_{a,k}-V_{a,c,k}\) the airspeed error. The terms \(\sigma_k\) and \(\rho_{n,k}\) penalize saturation and load risk, the weights \(w_r,w_h,w_V,w_u,w_s,w_n\) trade these objectives against one another, and \(\gamma\in(0,1]\) discounts future stages. Equation~\eqref{eq:objective} expresses the design pressure that the residual layer balances; the supervisor that resolves it is specified in the next section.


\section{Proposed Q-learning strategy}
\label{sec:method}

\subsection{The proposed structure}
\begin{figure}[htbp]
\centering
\includegraphics[width=\textwidth,height=0.70\textheight,keepaspectratio]{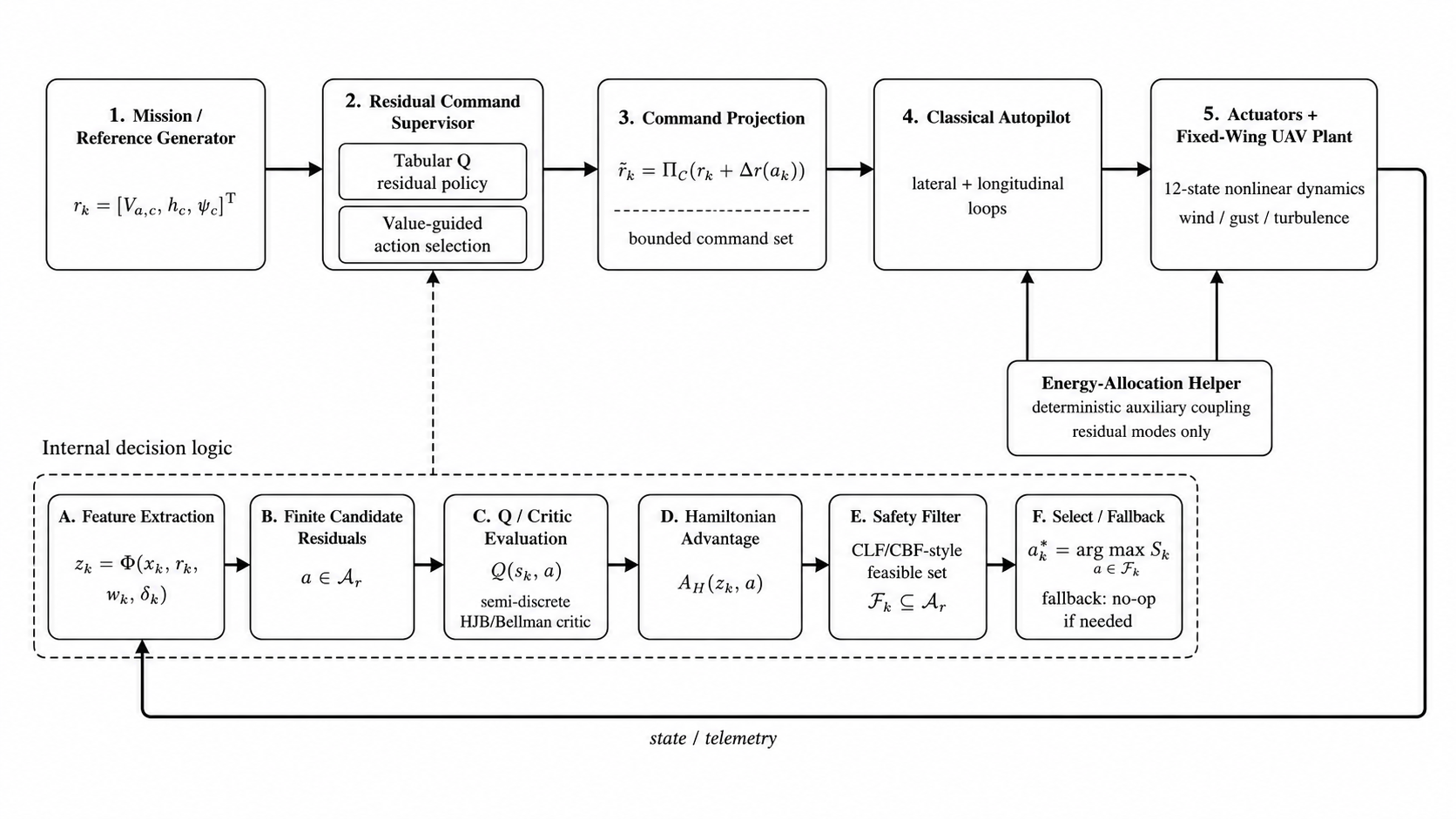}
\caption{Layered command-supervision framework shared by all controller modes.  The mission reference passes through an optional residual supervisor and the command-projection operator into the fixed gain-scheduled autopilot, the actuator model, and the plant.  The learned intervention is the bounded command residual \(\Delta r(a_k)\) only; the autopilot is the sole actuator-facing controller.}
\label{fig:framework_architecture}
\end{figure}

The residual supervisor reshapes the reference command and never touches the actuators directly.  Figure~\ref{fig:framework_architecture} shows the three roles that the framework keeps separate.  The main closed-loop path runs from the mission generator, which produces the nominal command \(r_k=[V_{a,c},h_c,\psi_c]^\top\), through an optional residual that modifies only that command, then through the projection operator that enforces the admissible command envelope, and finally into the fixed (gain-scheduled) autopilot that remains the only actuator-facing feedback controller.  Inside the supervisor, telemetry is encoded into features, seven bounded residual candidates are scored, tabular Q and a semi-discrete value-iteration term are combined, a finite-action shield removes hard-blocked candidates, and a no-op fallback stays available.  A separate telemetry and learning loop updates the reward, the Q values, the value-iteration diagnostics, and the logged shield flags.

Three source-supported choices shape this layout.  Learned authority sits at the command layer rather than the actuator layer.  The Beard--McLain monograph basis and gain-scheduled autopilot studies support preserving nested actuator-facing stabilization loops with explicit saturation and bandwidth handling \citep{beard_mclain_ngc,poksawat2017,liu2015}, while command-filtering and command-limiting work shows that modifying the reference can shape transients and reduce envelope exposure without replacing the lower-level controller \citep{dong2011,na2019,sun2021}.  The learned correction therefore stays confined to the residual \(\Delta r(a_k)\), not to elevator, aileron, rudder, or throttle.  The tabular-Q branch and the value-guided branch differ only in how they rank candidates; both pass through the same projection and the same classical autopilot.

The learning that we add is finite and auditable.  Residual RL motivates wrapping learned corrections around a nominal controller \citep{johannink2019,jayarathne2023}, and finite-abstraction work supports compact state and action summaries when interpretability and bounded computation matter more than high-dimensional approximation \citep{taherian2014}.  We follow that line with a seven-action residual set and a discretized state, so every candidate residual and its value are inspectable at each step.  The shield layer is a filter rather than a certificate.  Constraint-admissible supervision, CBF/CLF control, HJ safety analysis, and learned safe-value methods support action restriction and value-based risk reasoning \citep{li2018,ames2016,fisac2019,tan2024,zhao2023}, and we adapt that vocabulary to a finite residual set.  A deterministic energy-allocation helper may assist longitudinal energy management in the residual modes, but it is an auxiliary part of the controller package rather than a learned actuator policy.  The equations that follow define an auditable residual command supervisor with fallback behavior.

\FloatBarrier

All controller modes share the same plant, wind, actuator, and autopilot code path,
\begin{equation}
    x_{k+1}=F_{\Delta t}(x_k,\delta_k,w_k),
    \label{eq:shared_runtime}
\end{equation}
where \(F_{\Delta t}\) is the RK4 plant update and \(w_k\) denotes environmental disturbance inputs.  The command entering the autopilot is
\begin{equation}
\tilde r_k=
\begin{cases}
 r_k, & \text{baseline},\\
 \Pi_{\mathcal C}(r_k+\Delta r(a_k^{Q})), & \text{tabular Q residual},\\
 \Pi_{\mathcal C}(r_k+\Delta r(a_k^{H})), & \text{\method}.
\end{cases}
\label{eq:method_modes}
\end{equation}
The actuator-facing command is then
\begin{align}
\delta^{\mathrm{AP}}_{c,k} &= \kappa_{\mathrm{AP}}(x_k,\tilde r_k), \label{eq:ap_command}\\
\delta_{c,k} &=
\begin{cases}
\delta^{\mathrm{AP}}_{c,k}, & \text{baseline},\\
\mathcal E_{\mathrm{eng}}(\delta^{\mathrm{AP}}_{c,k},x_k,\tilde r_k,w_k), & \text{residual modes when the energy-allocation helper is active},\\
\delta^{\mathrm{AP}}_{c,k}, & \text{residual modes otherwise},
\end{cases} \label{eq:energy_helper_map}\\
\delta_k &= \mathcal D_{\mathrm{act}}(\delta_{c,k}), \label{eq:actuator_command}
\end{align}
where \(\kappa_{\mathrm{AP}}\) is the baseline autopilot, \(\mathcal D_{\mathrm{act}}\) is the actuator lag/rate/saturation operator, and \(\mathcal E_{\mathrm{eng}}\) is the residual-mode energy-allocation helper.  Because the plant, disturbances, actuator dynamics, autopilot, and metrics are common across modes, the benchmark compares controller packages whose only difference is the bounded command residual and its energy-allocation helper.  The implementation modules behind each operator are listed in the Code availability statement.

\subsection{Baseline autopilot and energy-allocation helper}
The baseline autopilot follows the successive-loop-closure small-UAV control architecture of the Beard--McLain monograph \citep{beard_mclain_ngc}.  Its command input is \(\tilde r_k=[V_{a,c,k},h_{c,k},\psi_{c,k}^{\circ}]^\top\), with the heading entry stored in degrees.  The lateral loops regulate yaw heading through commanded roll, roll-rate damping, and a sideslip rudder loop; the longitudinal loops use a state machine to select takeoff, climb, descend, or altitude-hold behavior.  In compact form,
\begin{equation}
    \delta_c=\kappa_{\mathrm{AP}}(x_k,\tilde r_k,\theta_{\mathrm{AP}}),
    \label{eq:autopilot_map}
\end{equation}
where \(\theta_{\mathrm{AP}}\) denotes the fixed (gain-scheduled) gains and \(\delta_c=[\delta_{e,c},\delta_{a,c},\delta_{r,c},\delta_{t,c}]^\top\) contains elevator, aileron, rudder, and throttle commands before the downstream actuator operator.  The controller tracks yaw heading \(\psi\), not ground-track course \(\chi\).  This is a project-specific simplification relative to path-following formulations under wind, where small-UAV guidance separates air-relative from inertial quantities and low-altitude wind-shear path-following often adds ground-velocity estimation or vector-field logic \citep{beard_mclain_ngc,zhang2024}.  We state the simplification explicitly because wind can separate heading from ground track.  The heading error is wrapped as
\begin{equation}
    e_{\psi,k}=\operatorname{wrap}_{[-\pi,\pi]}(\psi_{c,k}-\psi_k),
    \label{eq:heading_error}
\end{equation}
which avoids a discontinuity at \(\pm\pi\).

The heading and roll loops are
\begin{align}
\phi_{c,k} &= \operatorname{sat}_{[-45^\circ,45^\circ]}
\left(k_{p\psi}e_{\psi,k}+k_{i\psi}\sum_{j\le k}e_{\psi,j}T_s-k_{d\psi}r_{\mathrm{yaw},k}\right), \label{eq:heading_loop}\\
\delta_{a,c,k} &= \operatorname{sat}_{[-45^\circ,45^\circ]}
\left(k_{p\phi}(\phi_{c,k}-\phi_k)+k_{i\phi}\sum_{j\le k}(\phi_{c,j}-\phi_j)T_s-k_{d\phi}p_k\right). \label{eq:roll_loop}
\end{align}
Here \(r_{\mathrm{yaw},k}\) denotes the body yaw rate, not the command vector \(r_k=[V_{a,c,k},h_{c,k},\psi_{c,k}]^\top\).
The pitch loop is
\begin{equation}
\delta_{e,c,k}=\operatorname{sat}_{[-45^\circ,45^\circ]}
\left(\delta_{e}^{\ast}+k_{p\theta}(\theta_{c,k}-\theta_k)+k_{i\theta}\sum_{j\le k}(\theta_{c,j}-\theta_j)T_s-k_{d\theta}q_k\right),
\label{eq:pitch_loop}
\end{equation}
where \(\delta_e^{\ast}\) is the trim elevator component.  Airspeed regulation with throttle in altitude-hold mode is
\begin{equation}
\delta_{t,c,k}=\operatorname{sat}_{[0,1]}
\left(\delta_t^{\ast}+k_{pV}(V_{a,c,k}-V_{a,k})+k_{iV}\sum_{j\le k}(V_{a,c,j}-V_{a,j})T_s+k_{dV}\dot e_{V,k}\right).
\label{eq:airspeed_throttle_loop}
\end{equation}
The longitudinal state machine is
\begin{equation}
q^{AP}_k=\begin{cases}
1, & h_k\le h_{TO},\\
2, & h_k\le h_{c,k}-h_{hold},\\
3, & h_k\ge h_{c,k}+h_{hold},\\
4, & \text{otherwise},
\end{cases}
\label{eq:autopilot_state_machine}
\end{equation}
where \(h_{TO}\) is the takeoff-zone altitude threshold and \(h_{hold}\) is the altitude-hold deadband.  The four branches are runway/takeoff, climb/acceleration, descend, and altitude hold, with bumpless-transfer logic to reduce pitch-command steps at transitions.  In the climb branch the throttle is held near a climb-biased value and \(\theta_c\) is limited by speed-error-dependent caps; in the descend branch the throttle is reduced and airspeed is regulated through pitch; in altitude hold the throttle regulates airspeed while commanded pitch regulates altitude.  These rules follow the energy-coupled logic of classical small-UAV autopilots, which balance speed, altitude, and actuator feasibility rather than minimize a single tracking metric.  The resulting actuator command is clipped, rate-limited, and lagged by the common actuator operator, so baseline and residual modes share the same downstream actuator dynamics.

A lightweight energy-allocation helper runs in the residual controller modes.  It activates when a residual is active or when the disturbance-energy summary \(D_k=\lVert w_k\rVert+2\sigma_{\mathrm{turb},k}\) reaches the threshold \(D_k\ge4.0\).  The helper computes total and balance energy errors
\begin{align}
e_{T,k} &= g(h_{c,k}-h_k)+\frac{1}{2}(V_{a,c,k}^{2}-V_{a,k}^{2}), \label{eq:energy_total}\\
e_{B,k} &= g(h_{c,k}-h_k)-\frac{1}{2}(V_{a,c,k}^{2}-V_{a,k}^{2}), \label{eq:energy_balance}
\end{align}
and applies clipped throttle, pitch-command, and small elevator-bias assistance before the actuator dynamics.  The throttle term can raise the autopilot throttle command, the pitch term can override commanded pitch outside takeoff, and the elevator bias is added before the actuator operator.  The helper is not a full total-energy-control design and not an adaptive controller.  Because it belongs to the residual-mode package, the reported improvements are system-level evidence for the residual-supervision architecture and not an ablation-isolated effect of the value-iteration term alone.

\subsection{Finite residual command interface}
The residual action set is finite and shared by both supervisors, and the same projection is applied before commands reach the autopilot.  The set is
\begin{equation}
\mathcal A_r=\{a_0,a_{V+},a_{V-},a_{h+},a_{h-},a_{\psi+},a_{\psi-}\},
\label{eq:action_set}
\end{equation}
with no-op
\begin{equation}
\Delta r(a_0)=\begin{bmatrix}0&0&0\end{bmatrix}^{\top}.
\label{eq:noop}
\end{equation}
Each residual is a small command increment, not an actuator increment.  Table~\ref{tab:residual_action_table} lists the seven increments and the projection bounds.

\begin{table}[htbp]
\centering
\caption{Finite residual action set and command-projection bounds.}
\label{tab:residual_action_table}
\footnotesize
\begin{tabularx}{\textwidth}{P{0.18\textwidth}P{0.30\textwidth}Y}
\toprule
Action & Residual increment & Function in the supervisor \\
\midrule
\(a_0\) & \([0,0,0]^\top\) & no-op, preserve nominal mission command \\
\(a_{V+}\) & \([+2~\mathrm{m/s},0,0]^\top\) & request more speed/energy margin \\
\(a_{V-}\) & \([-2~\mathrm{m/s},0,0]^\top\) & reduce commanded airspeed when excess speed is present \\
\(a_{h+}\) & \([0,+10~\mathrm{m},0]^\top\) & bias toward climb/altitude recovery \\
\(a_{h-}\) & \([0,-10~\mathrm{m},0]^\top\) & reduce altitude command for energy relief or overshoot recovery \\
\(a_{\psi+}\) & \([0,0,+3^\circ]^\top\) & small positive heading correction \\
\(a_{\psi-}\) & \([0,0,-3^\circ]^\top\) & small negative heading correction \\
Projection & \(V_a\in[20,140]~\mathrm{m/s}\), \(h\in[0,450]~\mathrm{m}\), heading wrapped & command-interface admissibility \\
\bottomrule
\end{tabularx}
\end{table}

\subsection{Tabular Q residual supervisor}
The Q residual supervisor maps telemetry into a finite state abstraction
\begin{equation}
    s_k=\eta(e_{V,k},e_{h,k},e_{\mathrm{ref},k},e_{\perp,k},e_{r,k},V_{w,k},\sigma_{\mathrm{turb},k})\in\mathcal S_Q,
    \label{eq:q_state}
\end{equation}
where \(e_{\perp,k}\) is cross-track error for straight legs, \(e_{r,k}\) is radial error for orbit-like references, \(V_{w,k}\) is the wind-vector norm, and \(\sigma_{\mathrm{turb},k}\) is the turbulence standard-deviation setting.  The encoder discretizes seven components: airspeed error, altitude error, reference error, cross-track error, radial error, wind/turbulence stress, and a low-energy flag.  This is a pragmatic finite abstraction for an auditable supervisor; its bins and hyperparameters remain part of the empirical design space \citep{taherian2014,eimer2023}.  The hard-condition score combines tracking, disturbance, saturation \(\sigma_k\), and post-step load-factor telemetry \(n_{z,k}\):
\begin{equation}
\begin{aligned}
\chi_k=\max\{&|e_{V,k}|/10,\ |e_{h,k}|/35,\ |e_{\mathrm{ref},k}|/75,\ |e_{\perp,k}|/20,\ |e_{r,k}|/35,\\
&|V_{w,k}|/8,\ |\sigma_{\mathrm{turb},k}|,\ \max(\sigma_k-0.65,0)/0.2,\ \max(|n_{z,k}|-3.5,0)/1.5\}.
\end{aligned}
\label{eq:hard_condition_score}
\end{equation}
If \(\chi_k<1\), the supervisor returns the no-op candidate set, which keeps the nominal command in calm flight.  The same no-op-only set is used for large calm-air horizontal reference error, treated as possible mission-geometry mismatch rather than a disturbance-rejection opportunity.  Otherwise the supervisor builds a finite valid subset that favors speed/energy recovery, altitude correction, or lateral correction depending on the error pattern and risk level.  Near high load or saturation risk, the valid set narrows to no-op, positive airspeed, and altitude-relief candidates.

The tabular Bellman update is
\begin{equation}
Q_{k+1}(s_k,a_k)=Q_k(s_k,a_k)+\alpha_Q\left[\rho_k+\gamma_Q\max_{a'\in\mathcal A_r}Q_k(s_{k+1},a')-Q_k(s_k,a_k)\right],
\label{eq:q_update}
\end{equation}
with \(\alpha_Q=0.12\), \(\gamma_Q=0.95\), and exploration rate \(\epsilon_Q=0.05\) decaying toward \(0.01\) when training is enabled.  The reward is
\begin{align}
\rho_k^{Q}=&-\left(\frac{|e_{V,k}|}{22}+\frac{|e_{h,k}|}{110}+\frac{|e_{\mathrm{ref},k}|}{150}+0.5\max(\sigma_k-0.75,0)+c_a(a_k)\right) \nonumber\\
&-0.25\rho_{risk,k}-2\mathbf 1_{\mathrm{viol},k},
\label{eq:q_reward}
\end{align}
with action cost
\begin{equation}
 c_a(a_k)=0.02\frac{|\Delta V_a(a_k)|}{2}+0.02\frac{|\Delta h(a_k)|}{10}+0.04\frac{|\Delta\psi(a_k)|}{3},
\label{eq:action_cost}
\end{equation}
and risk score
\begin{equation}
\rho_{risk,k}=\max\!\left(\frac{\max(|n_{z,k}|-3.5,0)}{2.5},\frac{\max(\sigma_k-0.70,0)}{0.28}\right).
\label{eq:risk_score}
\end{equation}
The violation indicator \(\mathbf 1_{\mathrm{viol},k}\) fires only after takeoff-like conditions, under high load factor or extreme saturation.  The reward exposes the trade-off among reference tracking, speed and altitude tracking, residual action cost, saturation, load risk, and violation events.  Logged Q diagnostics include reward, episode return, temporal-difference error, action index, exploration flag, state key, selected Q value, next-state maximum Q value, residual activity, hard-condition score, load factor, and risk score.
The \method\ mode keeps this reward structure but applies the HJB advantage adjustment of Algorithm~\ref{alg:supervisor} and a stronger risk penalty coefficient \(0.35\rho_{risk,k}\) in the shielded residual mode.

\subsection{Value-iteration critic and Hamiltonian advantage}
\method\ inherits the finite action set, state encoder, reward interface, and projection from the tabular Q supervisor and changes only the candidate-ranking stage.  It adds normalized features
\begin{equation}
    z_k=\Phi(x_k,r_k,w_k,\delta_k)\in\mathbb R^7,
    \label{eq:feature_vector}
\end{equation}
with the scaling
\begin{equation}
 z_k=\begin{bmatrix}
 e_{V,k}/18\\
 e_{h,k}/80\\
 e_{\mathrm{ref},k}/180\\
 e_{\ell,k}/80\\
 (V_{w,k}+2\sigma_{\mathrm{turb},k})/16\\
 \max(\sigma_k-0.55,0)/0.45\\
 \max(|n_{z,k}|-3,0)/3
 \end{bmatrix},
 \label{eq:feature_definition}
\end{equation}
where \(e_{\ell,k}\) is cross-track error on straight legs and radial error on orbit legs.  The value proxy blends a quadratic term with a discrete value-iteration term,
\begin{equation}
\widehat V_H(z)=(1-\lambda_d)z^\top P_Hz+\lambda_d\widehat V_d(g(z)),
\label{eq:hjb_value}
\end{equation}
where \(P_H\succeq0\) is a fixed diagonal weight, \(g(z)\) maps the continuous features to a small finite grid, and \(\widehat V_d\) is computed by value iteration over that grid. Concretely, \(g(z)\) assigns each of the seven normalized features to its nearest center in a fixed set---\(\{-1.5,0,1.5\}\) for the airspeed, altitude, and lateral-error features, \(\{0,0.7,1.8,4.0\}\) for the reference-error feature, \(\{0,0.8,1.6\}\) for the wind/turbulence-stress feature, and \(\{0,1\}\) for the saturation- and load-stress flags---so the value-iteration grid has \(3\times3\times4\times3\times3\times2\times2=1296\) cells; the tabular Q-table uses a separate, finer bin-edge encoder over the same signals.  The iteration is
\begin{equation*}
\widehat V_d^{(i+1)}(g)=\min_{a\in\mathcal A_r}\big[\ell_d(\Psi(g,a),a)+\gamma_Q\,\widehat V_d^{(i)}(g(\Psi(g,a)))\big],
\end{equation*}
initialized from the quadratic proxy and run for five sweeps, with the result cached and reused at runtime.  The blend uses \(\lambda_d=0.38\).  The grid object is a local critic for ranking seven residual commands, motivated by Hamiltonian-driven ADP, which treats Hamiltonian quantities as continuous-time temporal-difference objects and approximates HJB solutions under explicit assumptions \citep{yang2017,yang2021,yang2022,lutter2020}.

For a candidate action \(a\), the supervisor predicts a short-horizon feature transition
\begin{equation}
\hat z_{k+1}^{a}=\Psi(z_k,a),
\label{eq:local_predictor}
\end{equation}
using \(\Psi(z_k,a)\), a hand-coded and deliberately conservative one-step predictor that applies direction-dependent coupling rules to the un-normalized error and stress signals---for example, lowering the altitude command under an energy deficit is credited as energy recovery, while a heading residual raises the predicted load---and then renormalizes, passing the wind and turbulence stress features through unchanged. The predictor encodes qualitative directionality rather than a high-fidelity aerodynamic model. It then computes a composite next-state penalty
\begin{equation}
\ell_H(\hat z_{k+1}^{a},a)=\widehat V_H(\hat z_{k+1}^{a})+0.045\left(\frac{|\Delta V_a(a)|}{2}+\frac{|\Delta h(a)|}{10}+1.35\frac{|\Delta\psi(a)|}{3}\right).
\label{eq:hjb_stage_cost}
\end{equation}
Here \(\ell_H\) is not a classical instantaneous stage cost: it evaluates the value proxy at the predicted next state \(\hat z_{k+1}^{a}\) rather than at the current state. The heuristic candidate-ranking score is
\begin{equation}
\mathcal H(z_k,a)=\ell_H(\hat z_{k+1}^{a},a)+\gamma_H\widehat V_H(\hat z_{k+1}^{a})-(1-\beta_H)\widehat V_H(z_k),
\label{eq:hamiltonian_score}
\end{equation}
with \(\gamma_H=\gamma_Q=0.95\) and \(\beta_H=0.035\). Because the same predicted next-state value \(\widehat V_H(\hat z_{k+1}^{a})\) appears both inside \(\ell_H\) and in the explicit \(\gamma_H\widehat V_H(\hat z_{k+1}^{a})\) term, it enters \(\mathcal H\) with total weight \((1+\gamma_H)=1.95\); \(\mathcal H\) is therefore a heuristic ranking score that emphasizes the predicted next-state value, not a literal one-step Hamiltonian or discrete Bellman residual. The no-op-relative advantage is
\begin{equation}
A_H(z_k,a)=\mathcal H(z_k,a)-\mathcal H(z_k,a_0).
\label{eq:hamiltonian_advantage}
\end{equation}
A negative \(A_H\) marks a candidate as locally better than no-op under the proxy.

\subsection{Finite-action shield and no-op fallback}
The finite-action shield adapts the vocabulary of safe action filtering to the residual layer.  Where a CBF-QP enforces forward invariance over continuous controls and an HJ reachability computation builds a continuous safety value function, this filter evaluates only the finite residual candidates present at the current step \citep{li2018,ames2016,fisac2019}.  Let \(D_k=V_{w,k}+2\sigma_{\mathrm{turb},k}\), let \(\rho_{risk,k}\) be the risk score of Eq.~\eqref{eq:risk_score}, and let \(\Delta h(a)\) and \(\Delta\psi(a)\) be the altitude and heading components of the residual.  The hard block is
\begin{equation}
B_{\mathrm{hard}}(k,a)=\mathbf 1[e_{\mathrm{ref},k}>10\ \land\ D_k>10\ \land\ |\Delta\psi(a)|>0].
\label{eq:hard_block}
\end{equation}
The CLF-style value-growth test is
\begin{equation}
\mathcal G_{\mathrm{CLF}}(k,a)=\mathbf 1\left[\widehat V_H(\hat z_{k+1}^{a})\le (1+0.32)\widehat V_H(z_k)+0.08\max(1,\chi_k)\right].
\label{eq:clf_gate}
\end{equation}
The CBF-style runtime predicate is the finite-action rule
\begin{equation}
\begin{aligned}
\mathcal G_{\mathrm{CBF}}(k,a)=\mathbf 1[&\neg(e_{V,k}>10\ \land\ D_k>10\ \land\ \Delta h(a)>0)\\
&\land\ (\rho_{risk,k}\le0.65\ \lor\ a\in\{a_0,a_{V+},a_{h-}\})\\
&\land\ (\rho_{risk,k}\le0.95\ \lor\ a=a_0)].
\end{aligned}
\label{eq:cbf_gate}
\end{equation}
The filtered set is
\begin{equation}
\begin{aligned}
\mathcal F_k
={}&\{a\in\mathcal A_{\mathrm{valid}}(s_k):B_{\mathrm{hard}}(k,a)=0,\\
&\quad [(\mathcal G_{\mathrm{CLF}}(k,a)=1 \land
\mathcal G_{\mathrm{CBF}}(k,a)=1)\lor A_H(z_k,a)\le 0]\}\cup\{a_0\}.
\end{aligned}
\label{eq:filtered_set}
\end{equation}
A nonzero candidate is admitted when it passes both gates or when it improves the no-op-relative Hamiltonian score, unless a hard block is active.  No-op is added unconditionally, so the filtered set is never empty. The admissible set \(\mathcal A_{\mathrm{valid}}(s_k)\) from which the filter starts is itself state-dependent: it collapses to \(\{a_0\}\) when the hard-condition score is below its gate or when the reference error exceeds \(100\) m in calm air, and otherwise opens from \(\{a_0,a_{V+}\}\), admitting a speed-decrease, altitude-climb, altitude-relief, or single heading-correction action only when the corresponding error and energy conditions hold, and narrowing to \(\{a_0,a_{V+},a_{h-}\}\) near high risk (\(\rho_{risk,k}>0.65\)).
The combined score is
\begin{equation}
S_k(a)=Q(s_k,a)+\lambda_H[-A_H(z_k,a)]-0.01\ell_H(\hat z_{k+1}^{a},a)-\lambda_\rho\rho_{risk,k}(a)+b_{rec}(z_k,a),
\label{eq:combined_score}
\end{equation}
with \(\lambda_H=0.85\) and a small recovery bonus \(b_{rec}\) for high-reference-error or low-energy cases.  The selected action is
\begin{equation}
    a_k^H=\arg\max_{a\in\mathcal F_k}S_k(a).
    \label{eq:selected_action}
\end{equation}
If every nonzero residual is rejected, \(a_0\) remains in \(\mathcal F_k\) and the nominal command is preserved.

The value proxy does not dominate every regime, so the supervisor delegates to the plain tabular-Q policy under severe disturbance.  When \(D_k>10\) and the situation shows large reference error (\(e_{\mathrm{ref},k}>25\)), severe energy capture (\(e_{V,k}>8\) and \(e_{h,k}>80\) on a straight leg), or a near-capture condition (\(e_{\mathrm{ref},k}<25\), \(|e_{h,k}|<10\), and \(|e_{V,k}|<5\) under \(D_k>10\)), the value-scoring and shield stage is bypassed and the residual is taken from the tabular-Q supervisor of Eq.~\eqref{eq:q_update}.  This keeps behavior anchored to the simpler learned policy when the hand-designed feature predictor is least reliable.  Across the benchmark the supervisor logs the selected candidate's value, advantage, stage cost, whether any candidate was removed by the shield, the surviving candidate count, residual activity, load factor, and risk score.

\begin{figure}[htbp]
\centering
\includegraphics[width=\textwidth,height=0.70\textheight,keepaspectratio]{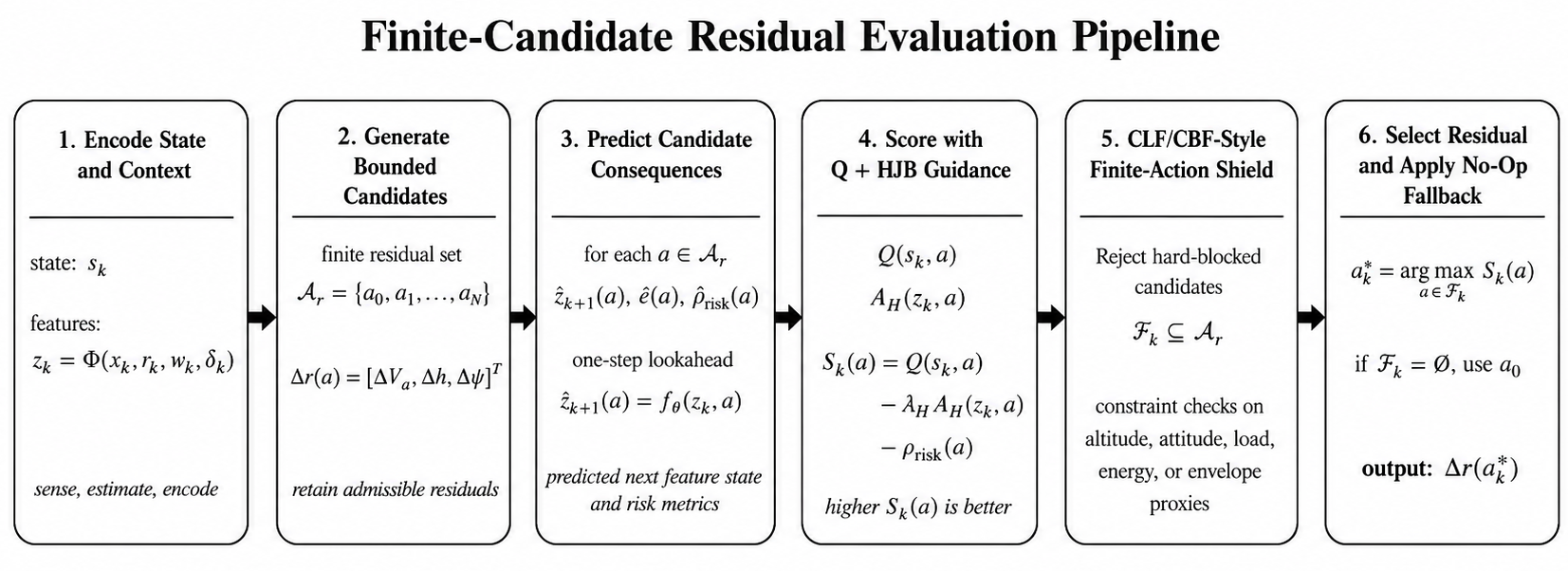}
\caption{HJB residual finite-candidate pipeline.  The supervisor encodes state and context, enumerates the seven bounded residual candidates \(\mathcal A_r=\{a_0,\ldots,a_6\}\), predicts one-step feature consequences, scores each candidate with the tabular Q term and the value-iteration guidance term, applies the finite-action shield, and dispatches the selected residual with no-op fallback.}
\label{fig:pipeline}
\end{figure}
\FloatBarrier

\subsection{Online learning loop}
Algorithm~\ref{alg:supervisor} states the runtime loop and Figure~\ref{fig:pipeline} is its graphical counterpart.  The loop computes errors and risk summaries from telemetry, gates learning out of nominal flight through the hard-condition score, evaluates every finite residual, selects only among candidates not removed by the shield while keeping no-op available, and updates the tabular critic when training is enabled.  Under severe disturbance it delegates the residual decision to the tabular-Q supervisor.  Evaluating the full candidate list once per step gives candidate-ranking complexity \(O(|\mathcal A_r|)\) for a fixed feature dimension and value table.

\textbf{Training and evaluation status.} The value-iteration table \(\widehat V_d\) is computed once at the start of a run, by five Bellman sweeps on the 1296-cell grid, and is then cached and reused unchanged. The tabular Q-values of both residual supervisors, by contrast, are updated online during the benchmark episodes: no separate pre-training phase precedes the benchmark, and the benchmark scenarios serve simultaneously as the learning environment, so the reported residual-method metrics reflect online adaptation rather than held-out evaluation of a frozen policy. The baseline receives no online updates. This is a deliberate choice that lets the residual supervisors adapt to each scenario's disturbance profile, but it means the residual-method numbers conflate within-run adaptation with tracking quality.

\begin{algorithm}[H]
\caption{HJB residual command supervision loop}
\label{alg:supervisor}
\begin{algorithmic}[1]
\Require state $x_k$, nominal command $r_k$, wind/disturbance telemetry $w_k$, actuator/load telemetry, Q table $Q$, value proxy $\widehat V_H$, action set $\mathcal A_r$, command set $\mathcal C$
\Ensure projected command $\tilde r_k$, reward $\rho_k$, logged diagnostics
\State Compute $V_{a,k}$, $h_k$, $\psi_k$, $e_{V,k}$, $e_{h,k}$, $e_{\mathrm{ref},k}$, lateral error $e_{\ell,k}$, saturation ratio $\sigma_k$, and load factor $n_{z,k}$.
\State Encode $s_k\leftarrow\eta(\cdot)$ using Eq.~\eqref{eq:q_state}, encode $z_k\leftarrow\Phi(\cdot)$ using Eq.~\eqref{eq:feature_definition}.
\State Compute disturbance summary $D_k\leftarrow V_{w,k}+2\sigma_{\mathrm{turb},k}$.
\If{$D_k>10$ \textbf{and} ($e_{\mathrm{ref},k}>25$ \textbf{or} severe energy capture \textbf{or} near-capture condition)}
    \State Take $a_k^H$ from the tabular-Q supervisor (Eq.~\eqref{eq:q_update}); \textbf{go to} command projection.
\EndIf
\State Compute hard-condition score $\chi_k$ using Eq.~\eqref{eq:hard_condition_score}.
\If{$\chi_k<1$}
    \State Set $\mathcal A_{\mathrm{valid}}\leftarrow\{a_0\}$.
\Else
    \State Build $\mathcal A_{\mathrm{valid}}\subseteq\mathcal A_r$ from energy, altitude, lateral, wind, saturation, and load-risk gates.
\EndIf
\State Compute no-op prediction $\hat z_{k+1}^{a_0}$, $\mathcal H(z_k,a_0)$, and $\widehat V_H(\hat z_{k+1}^{a_0})$.
\For{$a\in\mathcal A_{\mathrm{valid}}$}
    \State Predict $\hat z_{k+1}^{a}\leftarrow\Psi(z_k,a)$.
    \State Evaluate $\widehat V_H(\hat z_{k+1}^{a})$, $\ell_H(\hat z_{k+1}^{a},a)$, $\mathcal H(z_k,a)$, and $A_H(z_k,a)$.
    \State Evaluate CLF-style value-growth test Eq.~\eqref{eq:clf_gate} and CBF-style risk test Eq.~\eqref{eq:cbf_gate}.
    \State Mark $a$ as shielded if a hard block applies, or if it fails the CLF/CBF-style gates and does not improve the Hamiltonian relative to no-op.
    \State Compute $S_k(a)$ using Eq.~\eqref{eq:combined_score}.
\EndFor
\State Form $\mathcal F_k$ using Eq.~\eqref{eq:filtered_set}, include $a_0$ unconditionally.
\State Select $a_k^H\leftarrow\arg\max_{a\in\mathcal F_k}S_k(a)$.
\State Project $\tilde r_k\leftarrow\Pi_{\mathcal C}(r_k+\Delta r(a_k^H))$.
\State Send $\tilde r_k$ to the fixed (gain-scheduled) autopilot, execute plant/actuator step.
\State Compute reward using the Eq.~\eqref{eq:q_reward} structure with the \method\ risk coefficient and HJB reward adjustment $0.04\max(0,-A_H)-0.02\max(0,A_H)$.
\State Update $Q$ by Eq.~\eqref{eq:q_update}, log reward, TD error, residual activity, candidate count, shield status, HJB value, HJB advantage, saturation, load factor, and violation flag.
\end{algorithmic}
\end{algorithm}

\subsection{Structural properties}
Four properties follow from the design.  Every projected command satisfies \(\tilde r_k\in\mathcal C\) because projection is applied after the residual is added.  The baseline command stays available because \(a_0\in\mathcal A_r\) and is retained unconditionally in \(\mathcal F_k\).  The online candidate evaluation is finite because \(|\mathcal A_r|=7\).  All comparisons run in a shared runtime: Eq.~\eqref{eq:method_modes} captures the command-layer difference before the common autopilot, and the residual modes also pass through the energy-allocation helper before the common actuator and plant update.  Table~\ref{tab:method_comparison} summarizes how the three modes differ in authority and decision logic.

\begin{table}[htbp]
\centering
\caption{Controller modes compared by command authority and decision logic.}
\label{tab:method_comparison}
\footnotesize
\begin{tabularx}{\textwidth}{P{0.20\textwidth}P{0.34\textwidth}Y}
\toprule
Method & Authority and interface & Decision logic \\
\midrule
Baseline autopilot \citep{beard_mclain_ngc,poksawat2017,liu2015} & Only the classical autopilot commands actuators; no learned residual is present. & Fixed (gain-scheduled) lateral and longitudinal loops with actuator saturation and rate logic. \\
Tabular Q residual \citep{johannink2019,taherian2014,zahmatkesh2022} & Autopilot remains the actuator-facing controller; Q modifies only $V_{a,c}$, $h_c$, and $\psi_c$, with energy-allocation helper support when active. & Finite tabular Q-learning over discretized tracking, wind, lateral/path, and risk features. \\
\method\ \citep{fisac2019,yang2017,yang2021,yang2022,ames2016,li2018} & Autopilot remains the actuator-facing controller; value-guided logic ranks finite command residuals through the same helper path. & Q score plus semi-discrete value proxy, Hamiltonian advantage, finite-action shielding, no-op fallback, and tabular-Q delegation under severe disturbance. \\
\bottomrule
\end{tabularx}
\end{table}


\section{Experiment and results}
\label{sec:protocol}

We ran every test in one shared software-in-the-loop simulation runtime and compared three controller packages inside it. The first package is the classical autopilot alone, which sends the mission reference \(r_k\) directly to the inner loop. The second adds a tabular-Q residual supervisor that selects a bounded residual command and projects it back into the admissible envelope. The third, \method, uses the same residual interface but ranks candidates with an HJB/Bellman-inspired value, a no-op-relative Hamiltonian advantage, and a CLF/CBF-style filter, and keeps the no-op action as an always-available fallback. The three packages share the plant, autopilot, and actuator model and are compared as whole controller packages. They share the nonlinear 12-state plant, the wind-relative forces and moments, the altitude-dependent density model, the actuator lag/rate/saturation model, the classical autopilot, the mission generator, the wind, gust, and turbulence inputs, the integration step, and the metric definitions. Holding everything but the decision layer fixed is what lets us attribute the differences we report to the controller package rather than to the simulation.

The two residual packages also activate the energy-allocation helper of Section~\ref{sec:method} when residual activity or disturbance energy is high, whereas the baseline does not. Because that helper is part of every residual mode, the difference between baseline and residual packages reflects the residual-decision layer together with this helper, so the comparison is at the level of the controller package as a whole.

The primary evidence is a full-duration 20-scenario benchmark. Each controller package runs one full-duration episode per scenario, for 60 controller-scenario runs in total. Scenario durations range from 45 to 60~s, and seeds run from 4101 to 4120, one seed per scenario. The scenarios span loiter and tight orbits, racetrack and figure-eight references, straight climbs, runway-takeoff acceleration and climbout cases, and fight-mode S-turns. The disturbance fields combine steady north, east, and down wind components with body-axis gusts, turbulence, crosswind, headwind, tailwind, and mixed diagonal stressors. Appendix~\ref{app:scenario_catalog} lists the geometry and disturbance values for each scenario.

A coarse \(20\times 50\) sweep provides short-horizon evidence. It evaluates the three packages over the 20 scenarios with 50 repetitions each, for 3000 early-phase episodes. Because these episodes are early-phase rather than full missions, the sweep examines short-horizon seed and repetition sensitivity; it is not merged with the primary benchmark. A single 60~s fight-mode run provides a longer-horizon check that the controllers remain numerically well-behaved. Together the full-duration benchmark, the coarse sweep, and the fight-mode run form three evidence tiers, ordered from full-mission tracking through short-horizon sensitivity to numerical sanity.

The evaluation metrics are scalar episode summaries. Let \(N\) be the number of finite runtime samples retained for an episode and \(T_s\) the runtime sample time. At sample \(k\) the runtime gives altitude \(h_k=-p_{d,k}\), altitude error \(e_{h,k}=h_{c,k}-h_k\), airspeed error \(e_{V,k}=V_{a,c,k}-V_{a,k}\), and horizontal reference error \(e_{\mathrm{ref},k}=\sqrt{(p_{n,k}-p^{\mathrm{ref}}_{n,k})^2+(p_{e,k}-p^{\mathrm{ref}}_{e,k})^2}\). The three tracking metrics are root-mean-square errors,
\begin{align}
\mathrm{RMS}_{\mathrm{ref}}&=\sqrt{\frac{1}{N}\sum_{k=1}^{N}e_{\mathrm{ref},k}^{2}}, \label{eq:metric_ref}\\
\mathrm{RMS}_{h}&=\sqrt{\frac{1}{N}\sum_{k=1}^{N}e_{h,k}^{2}}, \label{eq:metric_alt}\\
\mathrm{RMS}_{V}&=\sqrt{\frac{1}{N}\sum_{k=1}^{N}e_{V,k}^{2}}. \label{eq:metric_air}
\end{align}
The control-activity index is the integral of the squared actuator vector,
\begin{equation}
E_u=\sum_{k=1}^{N}\delta_k^\top\delta_k T_s,
\label{eq:metric_energy}
\end{equation}
where \(\delta_k\) is the four-channel actuator vector after lag, rate limiting, and saturation. Because \(\delta_k\) mixes control-surface deflections and throttle in runtime command units, \(E_u\) measures control activity rather than physical propulsion, battery, or fuel energy. The safety-violation fraction counts the share of airborne samples that exceed a load or saturation threshold,
\begin{equation}
f_{\mathrm{viol}}=\frac{1}{N}\sum_{k=1}^{N}\mathbf 1\big[t_k>2,\ h_k>5,\ (|n_{z,k}|>6\ \lor\ s_k>0.98)\big],
\label{eq:metric_violation}
\end{equation}
where \(s_k\) is the actuator saturation ratio. Table~\ref{tab:runtime_safety_flag} states the runtime flag. This fraction is an empirical simulator metric.

\begin{table}[H]
\centering
\caption{How the runtime decides a sample is a safety violation, and how those samples become an episode score.}
\label{tab:runtime_safety_flag}
\scriptsize
\begin{tabularx}{\textwidth}{P{0.22\textwidth}P{0.26\textwidth}Y}
\toprule
Runtime quantity & Threshold or logic & Role in the episode score \\
\midrule
Altitude \(h_k=-p_{d,k}\) & \(h_k>5\) m & First part of the airborne gate. \\
Runtime time after the step & \(t_k>2\) s & Second part of the airborne gate. \\
Load factor \(n_{z,k}\) & \(|n_{z,k}|>6\) & One violation source once the airborne gate holds. \\
Actuator saturation ratio \(s_k\) & \(s_k>0.98\) & Alternative violation source once the airborne gate holds. \\
Per-sample safety flag & \(\mathbf 1[(t_k>2)\land(h_k>5)\land((|n_{z,k}|>6)\lor(s_k>0.98))]\) & Counted once per retained sample. \\
Safety-time fraction & \(\big(\sum_k \mathbf 1[\cdot]\big)/\max(N,1)\) & Fraction of retained samples. \\
\bottomrule
\end{tabularx}
\end{table}

The peak load factor is
\begin{equation}
n_{z,\max}=\max_{1\le k\le N}|n_{z,k}|,
\label{eq:metric_load}
\end{equation}
with the per-sample load factor clipped to the runtime range,
\begin{equation}
 n_{z,k}=\mathrm{clip}(1-\dot{w}_k/g,-8,8).
\label{eq:metric_load_runtime}
\end{equation}
Residual and shield activity are sample-fraction counters,
\begin{equation}
f_{\mathrm{res}}=\frac{1}{N}\sum_{k=1}^{N}\mathbf 1[\mathrm{residual\_active}_k],\qquad
f_{\mathrm{sh}}=\frac{1}{N}\sum_{k=1}^{N}\mathbf 1[\mathrm{shield\_active}_k],
\label{eq:metric_activity}
\end{equation}
giving the fraction of samples with a nonzero residual and the fraction at which the finite-action shield removed a candidate (the shield-active counter logged by the runtime). The hard-condition score read by both residual supervisors is the max-normalized stress measure
\begin{equation}
\begin{aligned}
c_k=\max\{&|e_{V,k}|/10, |e_{h,k}|/35, |e_{\mathrm{ref},k}|/75, |e_{\mathrm{xt},k}|/20, |e_{\mathrm{rad},k}|/35, |w_k|/8, |\sigma_k|,\\
&\max(s_k-0.65,0)/0.2, \max(|n_{z,k}|-3.5,0)/1.5\}.
\end{aligned}
\label{eq:metric_hard_condition}
\end{equation}
The HJB value proxy and advantage are selected-candidate diagnostics reported by \method\ at each step and then averaged over the episode. For a selected candidate the runtime stores the predicted value \(\widehat V_{H,k}\) and the no-op-relative advantage \(\widehat A_{H,k}=H_{i,k}-H_{0,k}\), where \(H_{i,k}=g_{i,k}+\gamma \widehat V_{i,k}-(1-\beta)\widehat V_{\mathrm{cur},k}\). For the baseline and tabular-Q packages these diagnostics stay at their default values, since neither evaluates HJB candidates.

Table~\ref{tab:code_traced_metrics} collects each reported metric with its symbol, how it is computed, and the unit and reading it carries.

A methodological note on the altitude and airspeed columns: for the residual supervisors, \(h_c\) and \(V_{a,c}\) are the post-residual commands the supervisor dispatched to the autopilot, not the nominal mission commands, so these columns measure how well the autopilot tracked the supervisor's own (possibly energy-relief-adjusted) command, whereas the reference/path RMS column measures deviation from the nominal geometric mission path for every method. Altitude and airspeed error against the pre-residual nominal command are not logged, which limits reading the altitude and airspeed columns as a pure tracking-quality comparison against a fixed target.

\begin{table}[htbp]
\centering
\caption{The metrics we report: symbol, how each is computed, and how to read its value.}
\label{tab:code_traced_metrics}
\footnotesize
\setlength{\tabcolsep}{4pt}
\begin{tabularx}{\textwidth}{@{}lXX@{}}
\toprule
Metric and symbol & How it is computed & Unit and reading \\
\midrule
Reference RMS, \(\mathrm{RMS}_{\mathrm{ref}}\) & RMS of the horizontal distance to the nominal geometric mission path, \(\sqrt{\mathrm{mean}(e_{\mathrm{ref}}^2)}\); unaffected by the residual. & m; nominal mission-path tracking. \\
Altitude RMS, \(\mathrm{RMS}_{h}\) & RMS of \(h_c-h\), with \(h=-p_d\) and \(h_c\) the command actually sent to the autopilot (post-residual for the residual methods, nominal for the baseline). & m; tracking of the adjusted command, not the nominal mission altitude; the nominal-command altitude error is not separately logged. \\
Airspeed RMS, \(\mathrm{RMS}_{V}\) & RMS of \(V_{a,c}-V_a\), with \(V_{a,c}\) the command actually sent to the autopilot (post-residual for the residual methods, nominal for the baseline). & m\,s\(^{-1}\); tracking of the adjusted command; when a residual lowers \(V_{a,c}\) for energy recovery this metric reflects the reduced command. \\
Control activity, \(E_u\) & Integral \(\sum_k \delta_k^\top\delta_k T_s\) of the squared actuator vector. & actuator-command\(^2\,\)s; a runtime activity index, not physical propulsion or battery energy. \\
Safety-violation fraction, \(f_{\mathrm{viol}}\) & Fraction of samples with \(t>2\), \(h>5\), and either \(|n_z|>6\) or \(s>0.98\). & dimensionless fraction; a runtime threshold metric. \\
Max absolute load factor, \(n_{z,\max}\) & \(\max|n_z|\) of the clipped \(n_z=1-\dot{w}/g\). & g units; nonuniform and tie-heavy across methods. \\
Residual active fraction, \(f_{\mathrm{res}}\) & Fraction of samples with a nonzero residual action index. & dimensionless fraction; a residual-usage diagnostic. \\
Shield active fraction, \(f_{\mathrm{sh}}\) & Fraction of samples at which any candidate residual was shielded. & dimensionless fraction; a filter-activity diagnostic. \\
HJB value proxy, \(\bar V_H\) & Episode mean of the selected-candidate predicted value. & dimensionless normalized critic diagnostic. \\
HJB advantage, \(\bar A_H\) & Episode mean of the selected Hamiltonian difference relative to no-op. & dimensionless diagnostic; negative is favorable relative to no-op under the proxy. \\
Hard-condition score, \(\bar c\) & Episode mean of the max-normalized score of Eq.~\eqref{eq:metric_hard_condition}. & dimensionless stress gate, a heuristic runtime diagnostic. \\
\bottomrule
\end{tabularx}
\end{table}

These metrics make the comparison a package-level one. The stored episodes support claims about the baseline, tabular-Q, and \method\ packages under the shared runtime at the package level. The benchmark scores reference tracking together with the airspeed and energy cost of achieving it, because in these episodes better spatial tracking is not free: altitude and heading corrections draw on energy, speed margin, and actuator authority. The numerical claims throughout the paper derive from these simulated episodes; external literature motivates the modeling, control, residual-learning, safe-filtering, and HJB/ADP choices.

\subsection{Results}
\label{sec:results}

\begin{table}[htbp]
\centering
\caption{Full-duration 20-scenario aggregate metrics ($N=20$ episodes per method, one episode per scenario). Lower is better for every listed metric. Ctrl. idx. is the runtime control-activity index and Viol. frac. is the safety-violation fraction. Interval columns are across-scenario descriptive 95\% intervals in the same units as the adjacent metric (m, m/s).}
\label{tab:aggregate_results}
\footnotesize
\setlength{\tabcolsep}{3pt}
\begin{adjustbox}{max width=\textwidth}
\begin{tabular}{lrrrrrrrrrr}
\toprule
Method & Episodes & Ref./path RMS & Interval & Alt. RMS & Interval & Airspeed RMS & Interval & Ctrl. idx. & Viol. frac. & Max $|n_z|$\\
 & & (m) & (m) & (m) & (m) & (m/s) & (m/s) & & & \\
\midrule
Baseline & 20 & 338.617 & 129.248 & 114.137 & 11.312 & 10.546 & 0.626 & 5.665 & 0.004802 & 7.414\\
Q residual & 20 & 88.809 & 76.550 & 65.819 & 6.059 & 15.086 & 0.711 & 6.524 & 0.003090 & 6.844\\
HJB residual & 20 & 44.809 & 32.781 & 64.710 & 5.808 & 15.191 & 0.694 & 6.346 & 0.003332 & 6.800\\
\bottomrule
\end{tabular}
\end{adjustbox}
\end{table}

The full-duration 20-scenario aggregate in Table~\ref{tab:aggregate_results} is the primary result, computed from one full episode per scenario across all 20 scenarios. \method\ attains the lowest mean spatial reference/path RMS at 44.809~m, against 88.809~m for the Q residual package and 338.617~m for the baseline. \method\ also attains the lowest mean altitude RMS at 64.710~m, against 65.819~m for Q and 114.137~m for the baseline. The baseline attains the lowest mean airspeed RMS at 10.546~m/s, while Q reaches 15.086~m/s and \method\ reaches 15.191~m/s. The baseline also has the lowest mean control-activity index at 5.665, against 6.524 for Q and 6.346 for \method. Mean safety-violation fraction is 0.004802 for the baseline, 0.003090 for Q, and 0.003332 for \method, and mean maximum \(|n_z|\) is 7.414, 6.844, and 6.800 respectively.


\begin{table}[htbp]
\centering
\caption{Reduction in mean spatial reference/path RMS over the full-duration benchmark ($N=20$ episodes per method), in percent. Each value is computed from the aggregate means in Table~\ref{tab:aggregate_results}.}
\label{tab:improvements}
\begin{tabular}{lr}
\toprule
Comparison & Reduction in mean spatial reference/path RMS \\
\midrule
Q residual vs baseline & 73.77\%\\
HJB residual vs baseline & 86.77\%\\
HJB residual vs Q residual & 49.54\%\\
\bottomrule
\end{tabular}
\end{table}

Table~\ref{tab:improvements} converts the aggregate means of Table~\ref{tab:aggregate_results} into percentage reductions in mean spatial reference/path RMS. The Q residual package lowers the mean by 73.77\% relative to the baseline. \method\ lowers the mean by 86.77\% relative to the baseline and by 49.54\% relative to Q.

The reduction in the mean does not hold at the median. The median spatial reference/path RMS, computed from the full-precision per-scenario episode values, is 0.802~m for both Q and \method\ and 170.746~m for the baseline. Across the 20 scenarios, Q and \method\ produce equal medians, and the difference between their means comes from two high-disturbance scenarios in which Q diverges and \method\ recovers. In the fight crosswind-and-turbulence scenario, Q reaches 493.08~m while \method\ reaches 57.41~m; in the fight tailwind-and-lateral-gust scenario, Q reaches 620.02~m while \method\ reaches 214.96~m.


\begin{table}[htbp]
\centering
\caption{Per-scenario strict winner counts over the 20-scenario full-duration benchmark ($N=20$ scenarios). A method wins a scenario when it holds the unique minimum of the listed metric; exact ties are counted separately and listed in the tie column.}
\label{tab:winner_counts}
\begin{tabular}{lrrrr}
\toprule
Metric minimized & Baseline & Q residual & HJB residual & Tie note \\
\midrule
Spatial reference/path RMS & 4 & 1 & 15 & strict minima \\
Altitude RMS & 0 & 2 & 18 & strict minima \\
Airspeed RMS & 20 & 0 & 0 & strict minima \\
Control-activity index & 16 & 1 & 3 & strict minima \\
Safety-violation fraction & 3 & 0 & 1 & 16 exact ties \\
Max absolute load factor & 4 & 2 & 8 & 6 exact ties \\
\bottomrule
\end{tabular}
\end{table}

Table~\ref{tab:winner_counts} counts, metric by metric, how many of the 20 scenarios each method wins outright. For spatial reference/path RMS, \method\ wins 15 of 20 scenarios, the baseline wins 4, and Q wins 1. For altitude RMS, \method\ wins 18, Q wins 2, and the baseline wins 0. For airspeed RMS, the baseline wins all 20. For the control-activity index, the baseline wins 16, \method\ wins 3, and Q wins 1. Safety-violation fraction has 16 exact ties, with 3 baseline wins, 1 \method\ win, and 0 Q wins among the remainder; maximum absolute load factor has 6 exact ties, with 8 \method\ wins, 4 baseline wins, and 2 Q wins.

\begin{table}[htbp]
\centering
\caption{Full-duration aggregate metrics normalized to the baseline ($N=20$ episodes per method, dimensionless ratios). A ratio below one is lower than the baseline; a ratio above one is higher. Viol. frac. is the safety-violation fraction and Ctrl. idx. is the runtime control-activity index.}
\label{tab:normalized_tradeoffs}
\footnotesize
\setlength{\tabcolsep}{3pt}
\begin{adjustbox}{max width=\textwidth}
\begin{tabular}{lrrrrrr}
\toprule
Method & Ref./path RMS & Alt. RMS & Airspeed RMS & Ctrl. idx. & Viol. frac. & Max $|n_z|$\\
\midrule
Baseline & 1.000 & 1.000 & 1.000 & 1.000 & 1.000 & 1.000\\
Q residual & 0.262 & 0.577 & 1.431 & 1.152 & 0.643 & 0.923\\
HJB residual & 0.132 & 0.567 & 1.440 & 1.120 & 0.694 & 0.917\\
\bottomrule
\end{tabular}
\end{adjustbox}
\end{table}

Table~\ref{tab:normalized_tradeoffs} restates the same aggregate means as ratios to the baseline, so a value below one is lower than the baseline and a value above one is higher. Both residual packages are below one for spatial reference/path and altitude RMS and above one for airspeed RMS and the control-activity index. Airspeed RMS rises from 10.5~m/s under the baseline to 15.1~m/s under Q and 15.2~m/s under \method, and the control-activity index rises from 5.665 to 6.524 under Q and 6.346 under \method. No single package has the lowest value on every metric.

\begin{figure}[htbp]
\centering
\includegraphics[width=0.84\textwidth]{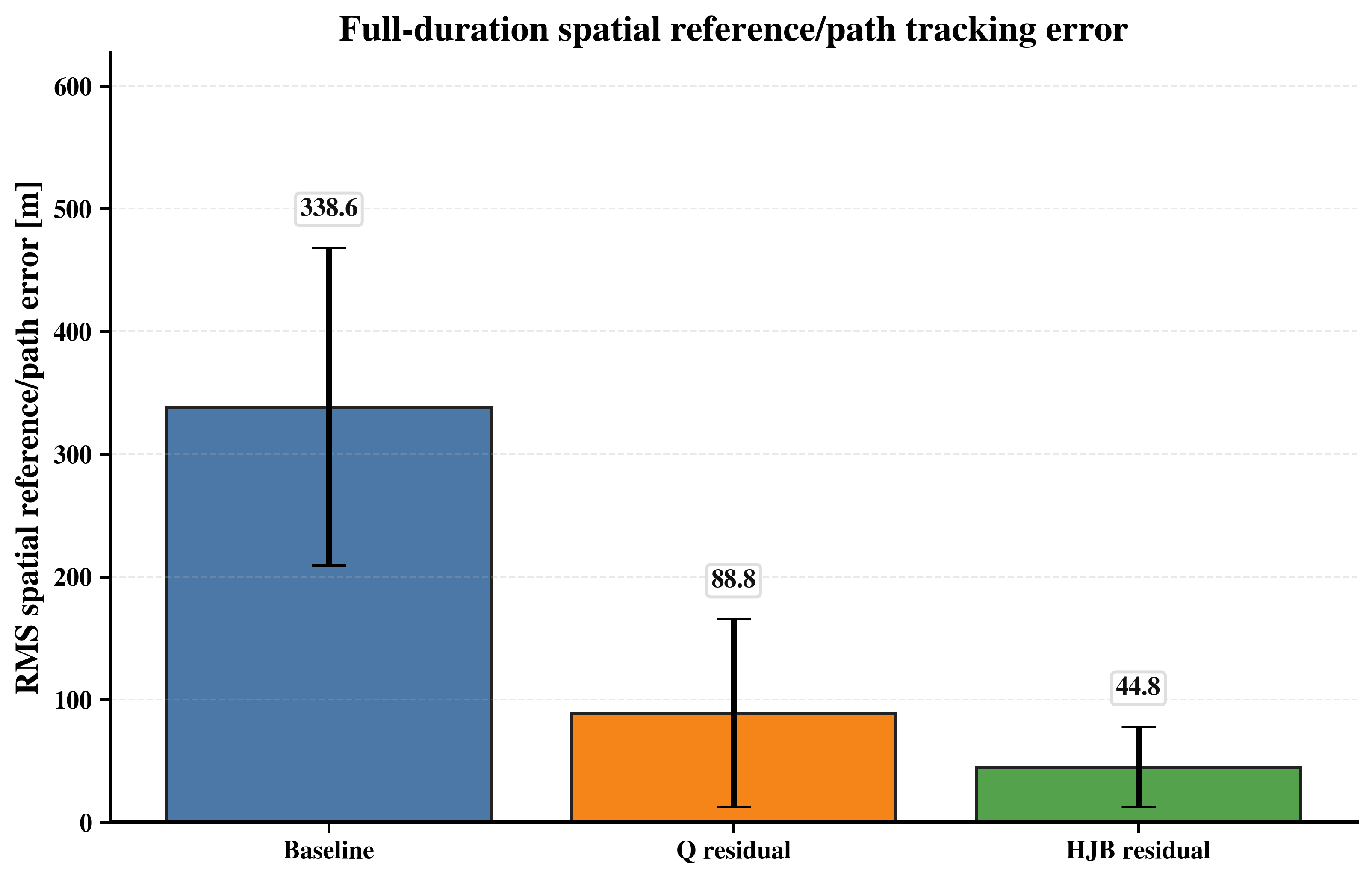}
\caption{Mean RMS spatial reference/path error over the 20-scenario full-duration benchmark ($N=20$ episodes per method, one episode per scenario), in metres. Lower is better. Bars rank baseline 338.617, Q residual 88.809, and \method\ 44.809. Error bars are across-scenario descriptive 95\% intervals.}
\label{fig:rms_reference}
\end{figure}
Figure~\ref{fig:rms_reference} plots the mean spatial reference/path RMS of Table~\ref{tab:aggregate_results} as a bar per method, with across-scenario descriptive 95\% intervals as error bars. The bars rank \method\ at 44.809~m below Q at 88.809~m and the baseline at 338.617~m, the 86.77\% and 49.54\% reductions reported above.

\begin{figure}[htbp]
\centering
\includegraphics[width=0.94\textwidth,height=0.70\textheight,keepaspectratio]{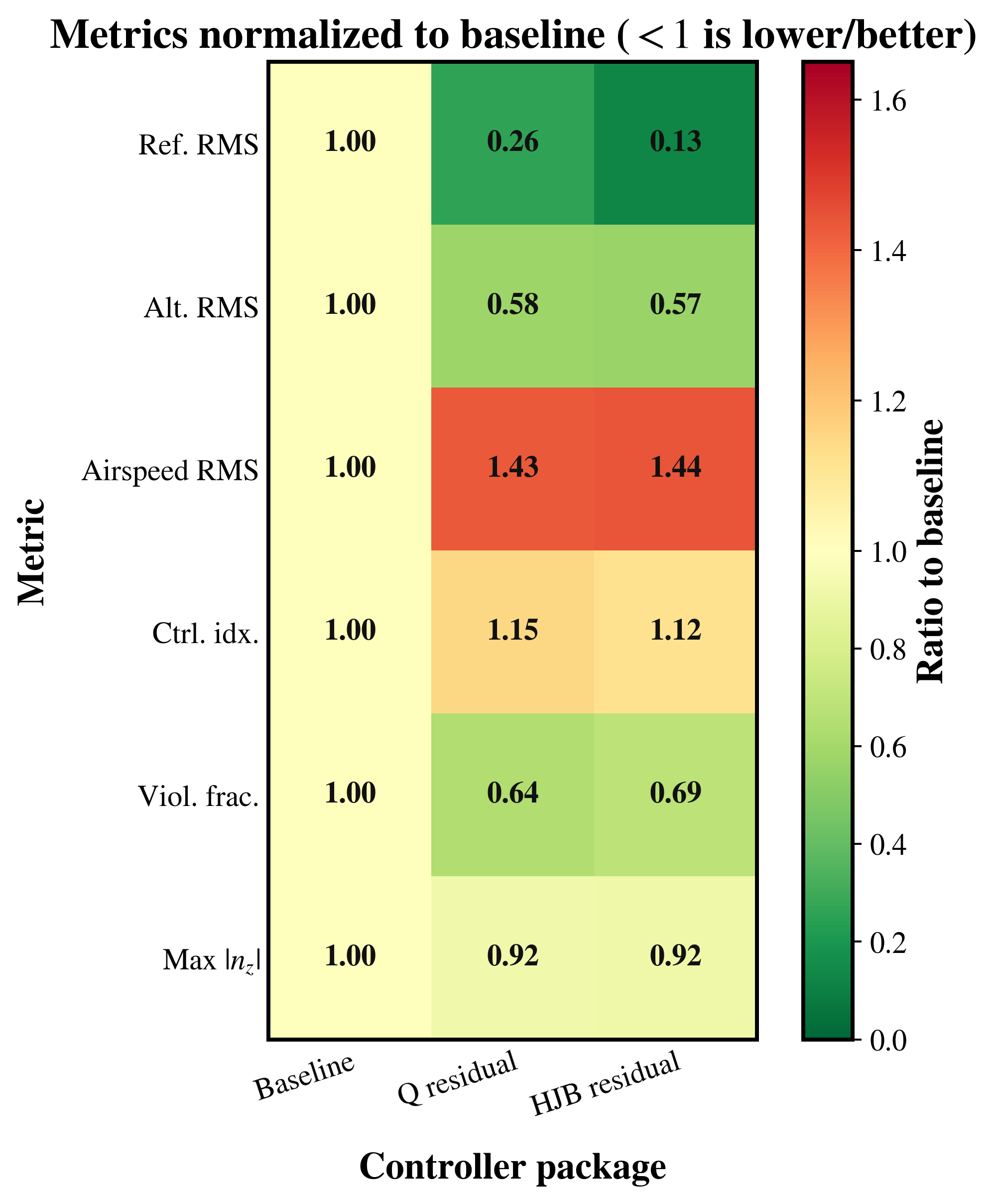}
\caption{Full-duration aggregate metrics normalized to the baseline ($N=20$ episodes per method). A ratio below one is lower than the baseline. \method\ is lowest for spatial reference/path and altitude error; both residual packages exceed the baseline for airspeed RMS and control-activity index.}
\label{fig:normalized_metrics}
\end{figure}
Figure~\ref{fig:normalized_metrics} draws the baseline-normalized ratios of Table~\ref{tab:normalized_tradeoffs} as a grouped bar chart. \method\ sits at 0.132 for spatial reference/path RMS and 0.567 for altitude RMS, both below one, while airspeed RMS reaches 1.440 and the control-activity index 1.120, both above one; the Q residual package follows the same below-and-above pattern at 0.262, 0.577, 1.431, and 1.152.

\begin{figure}[htbp]
\centering
\includegraphics[width=0.78\textwidth]{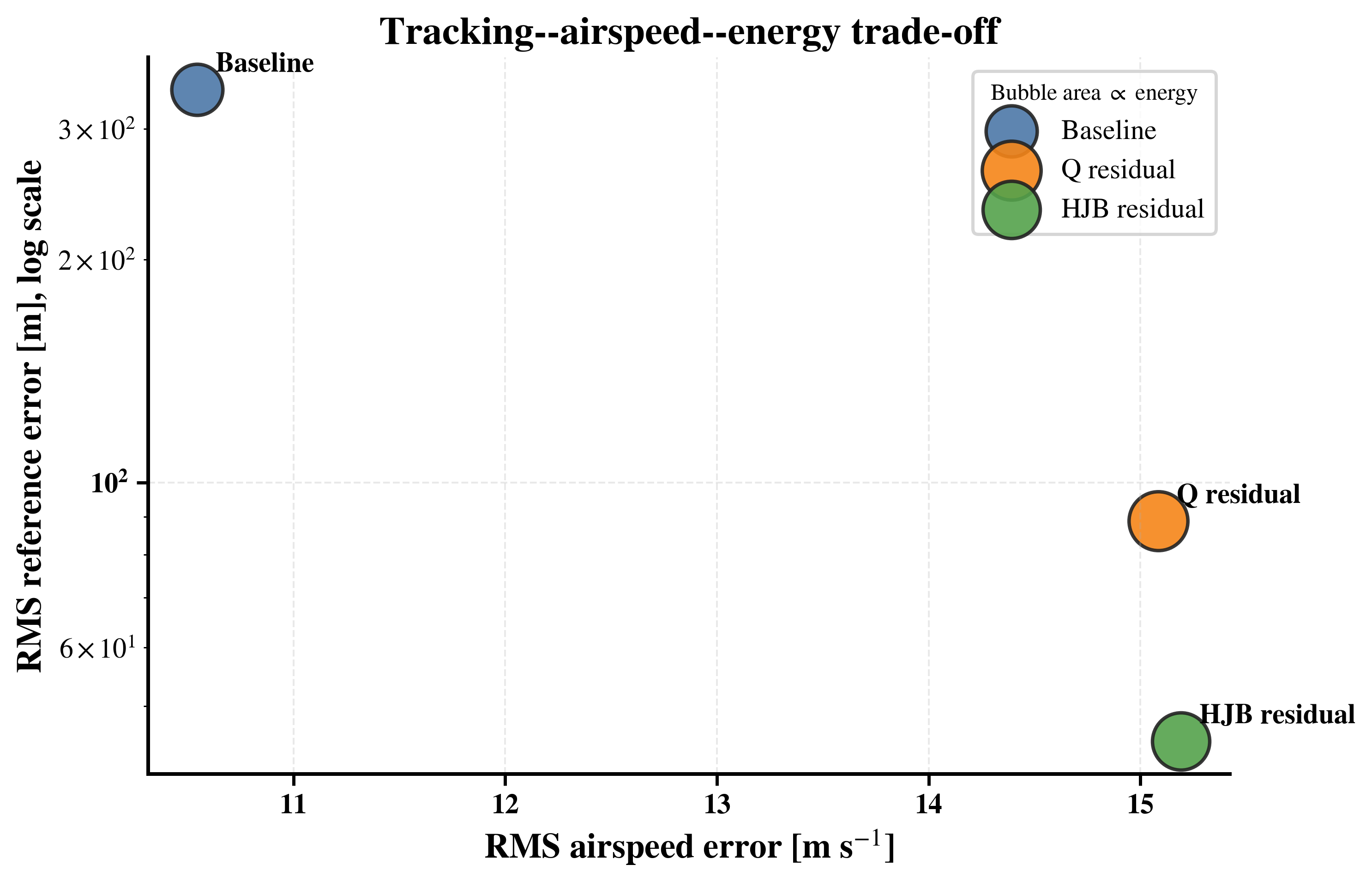}
\caption{Mean spatial reference/path RMS (m) versus mean airspeed RMS (m/s) over the full-duration benchmark ($N=20$ episodes per method); bubble area scales with the control-activity index. The baseline holds the lowest airspeed error, and the residual packages hold lower spatial reference/path error at higher airspeed error.}
\label{fig:tradeoff}
\end{figure}
Figure~\ref{fig:tradeoff} places each method on the spatial reference/path RMS axis against the airspeed RMS axis, with bubble area scaling with the control-activity index. The baseline sits at the lowest airspeed RMS of 10.546~m/s but the highest spatial reference/path RMS of 338.617~m, while \method\ moves to 44.809~m spatial RMS at 15.191~m/s airspeed RMS and Q to 88.809~m at 15.086~m/s, so the spatial gain comes alongside a higher airspeed error.

\begin{figure}[htbp]
\centering
\includegraphics[width=0.95\textwidth]{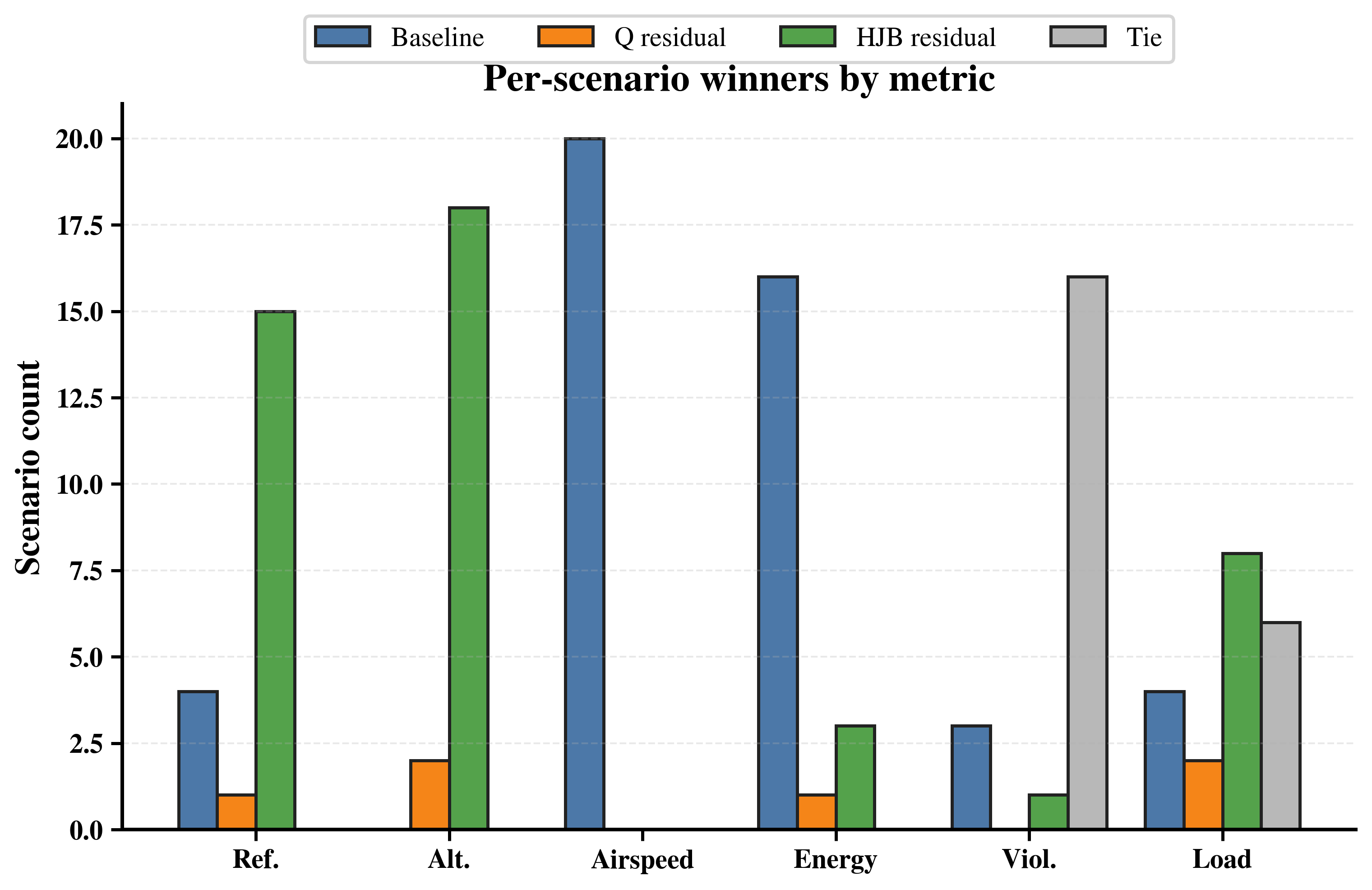}
\caption{Per-scenario strict winner counts over the 20-scenario full-duration benchmark ($N=20$ scenarios). Spatial reference/path and altitude wins favour the residual packages; airspeed and control-activity wins favour the baseline; safety-violation and load-factor metrics are tie-heavy.}
\label{fig:winner_counts}
\end{figure}
Figure~\ref{fig:winner_counts} shows the per-scenario winner counts of Table~\ref{tab:winner_counts} as stacked bars per metric. The spatial reference/path and altitude bars are dominated by \method\ at 15 and 18 wins, the airspeed bar is entirely baseline at 20 wins, the control-activity bar is mostly baseline at 16 wins, and the safety-violation and load-factor bars are short because 16 and 6 scenarios end in exact ties.

\begin{figure}[htbp]
\centering
\includegraphics[width=0.95\textwidth,height=0.72\textheight,keepaspectratio]{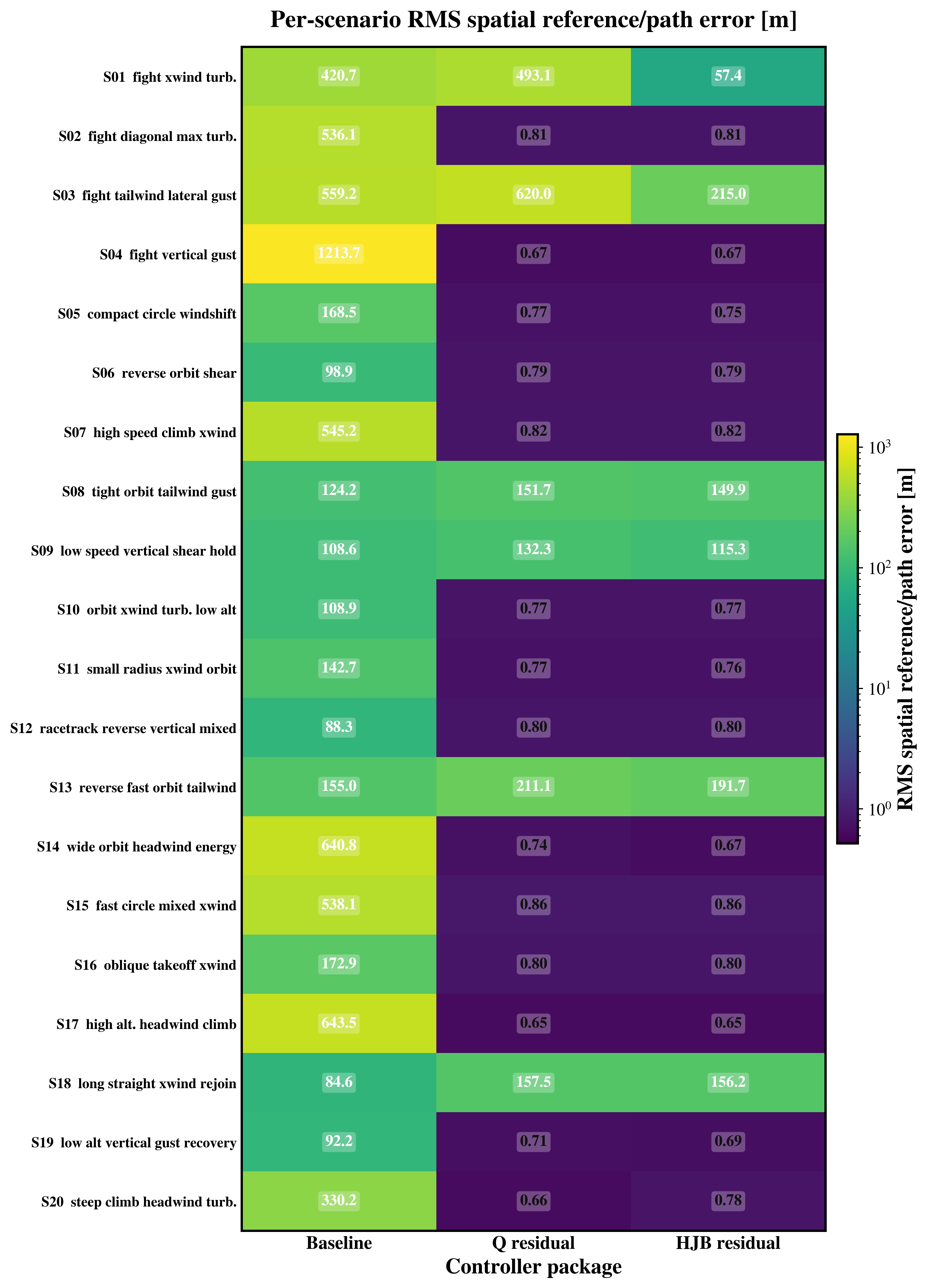}
\caption{Per-scenario spatial reference/path RMS (m) for the three methods over the full-duration benchmark ($N=20$ scenarios). Lower is better. Several high-disturbance scenarios carry large baseline errors that fall under residual supervision; a smaller set keeps the baseline or Q competitive.}
\label{fig:scenario_heatmap}
\end{figure}
Figure~\ref{fig:scenario_heatmap} lays out the spatial reference/path RMS of all 20 scenarios as a method-by-scenario heatmap. The baseline row carries the darkest cells, including the high-disturbance scenarios where its error reaches the hundreds of metres, and these fall under residual supervision, while a smaller set of scenarios keeps the baseline or Q cells comparable to \method, matching the 4 baseline and 1 Q spatial wins of Table~\ref{tab:winner_counts}.

\begin{table}[htbp]
\centering
\caption{Mean spatial reference/path RMS (m) by mission profile over the full-duration benchmark. Lower is better. The $N$ column gives the number of scenarios in each profile group, which differs across groups.}
\label{tab:profile_summary}
\footnotesize
\begin{tabularx}{\textwidth}{P{0.34\textwidth}rrrr}
\toprule
Profile & $N$ & Baseline & Q residual & HJB residual\\
\midrule
fight mode & 4 & 682.42 & 278.64 & 68.46\\
figure eight & 2 & 133.74 & 0.78 & 0.77\\
high speed climb s turn 200 & 2 & 334.68 & 76.25 & 75.38\\
loiter orbit & 3 & 120.07 & 44.60 & 38.94\\
racetrack & 3 & 294.70 & 70.87 & 64.41\\
runway takeoff accel 200 & 2 & 355.52 & 0.83 & 0.83\\
straight climb altitude hold & 2 & 364.06 & 79.06 & 78.42\\
takeoff climbout 200 & 2 & 211.18 & 0.68 & 0.74\\
\bottomrule
\end{tabularx}
\end{table}

Table~\ref{tab:profile_summary} groups mean spatial reference/path RMS by mission profile, averaging the scenarios in each group as counted in the $N$ column. The residual packages have lower mean spatial reference/path RMS than the baseline in the orbit, racetrack, climb, and fight-mode profile groups; in the fight-mode group the mean falls from 682.42~m for the baseline to 278.64~m for Q and 68.46~m for \method. In the takeoff-climbout group, Q reaches 0.68~m and \method\ reaches 0.74~m, so Q is marginally lower than \method\ there.

\begin{figure}[htbp]
\centering
\includegraphics[width=0.90\textwidth]{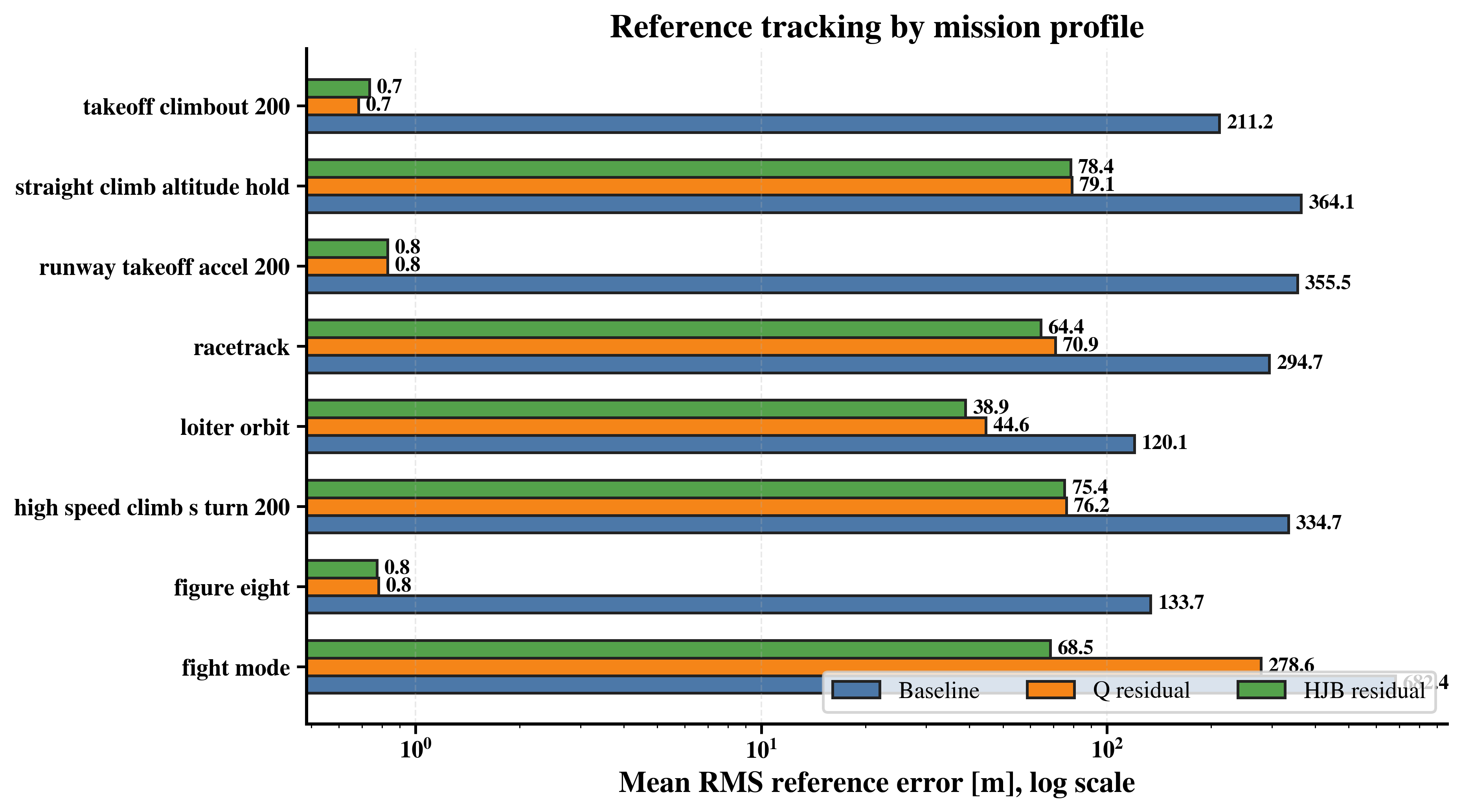}
\caption{Mean spatial reference/path RMS (m) by mission profile over the full-duration benchmark, with unequal scenario counts per profile group. Lower is better. The residual packages have lower mean spatial reference/path RMS than the baseline across several profile families.}
\label{fig:profile_summary}
\end{figure}
Figure~\ref{fig:profile_summary} draws the per-profile means of Table~\ref{tab:profile_summary} as grouped bars. The baseline bar is tallest in every profile group, the residual bars are far shorter in the fight-mode, racetrack, and climb groups, and the figure-eight, runway-takeoff, and takeoff-climbout groups collapse to under 1~m for both residual packages.


\begin{table}[htbp]
\centering
\caption{Mean residual-supervisor diagnostics over the full-duration benchmark ($N=20$ episodes per method). Residual-active and shield-active fractions are dimensionless; the hard-condition score, HJB value proxy, and HJB advantage are dimensionless surrogate quantities. The baseline and the Q residual package carry no HJB layer, so their HJB columns are zero.}
\label{tab:residual_diagnostics}
\small
\setlength{\tabcolsep}{3pt}
\begin{adjustbox}{max width=\textwidth}
\begin{tabular}{lrrrrr}
\toprule
Method & Residual active frac. & Hard-condition score & HJB value & HJB advantage & Shield active frac. \\
\midrule
Baseline & 0.000 & 0.000 & 0.000 & 0.000 & 0.000\\
Q residual & 0.913 & 2.959 & 0.000 & 0.000 & 0.000\\
HJB residual & 0.888 & 2.534 & 2.642 & -0.326 & 0.120\\
\bottomrule
\end{tabular}
\end{adjustbox}
\end{table}

Table~\ref{tab:residual_diagnostics} reports the mean residual-supervisor diagnostics logged during the full-duration benchmark. The Q residual package is active in a mean 0.913 fraction of the timeline and \method\ in 0.888, and \method\ logs shield activity in 0.120 of samples. The mean HJB advantage for \method\ is $-0.326$ and the mean HJB value proxy is 2.642. These quantities are zero for the baseline and for Q, which carry no HJB layer.

\begin{figure}[htbp]
\centering
\includegraphics[width=0.85\textwidth]{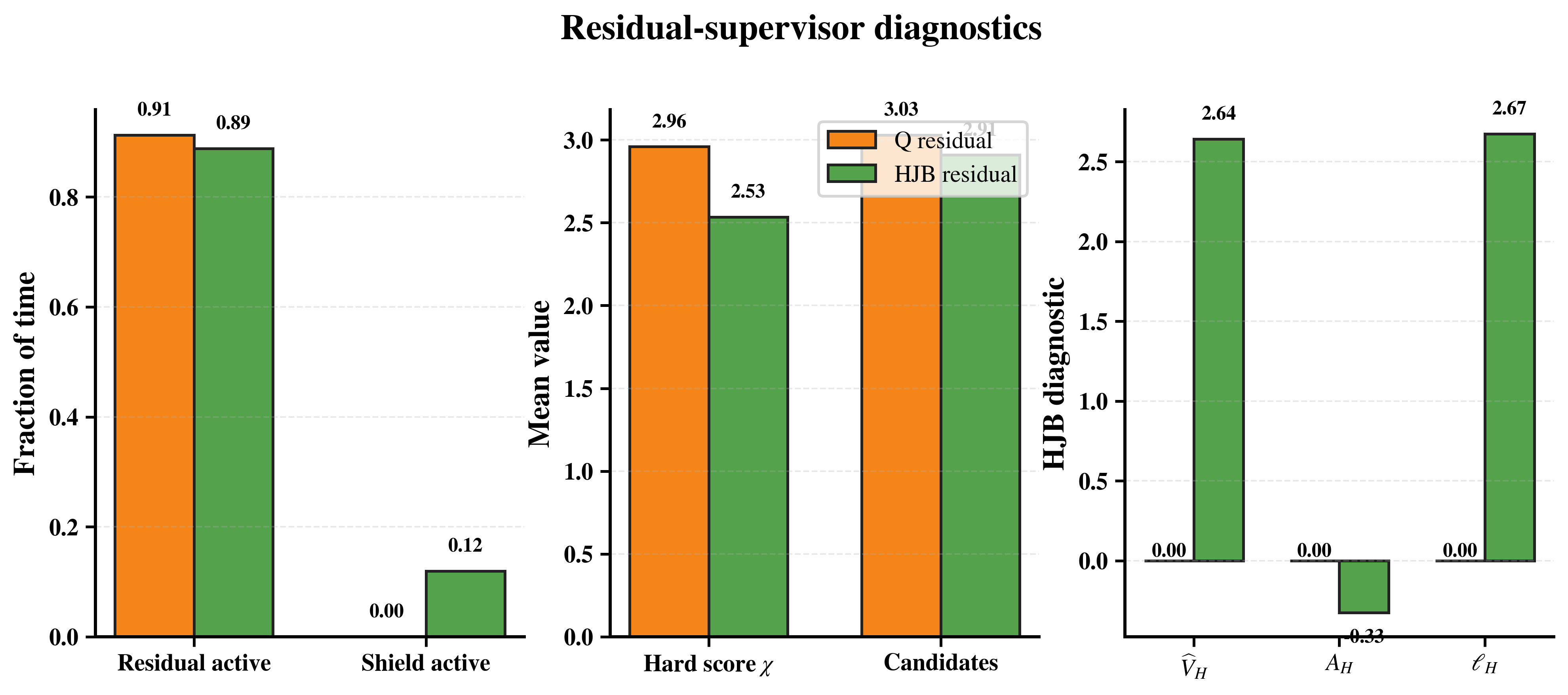}
\caption{Residual-supervisor diagnostics over the full-duration benchmark ($N=20$ episodes per method): residual-active fraction, hard-condition score, HJB value proxy, HJB advantage, and shield-active fraction. HJB columns are nonzero only for \method.}
\label{fig:residual_diagnostics}
\end{figure}
Figure~\ref{fig:residual_diagnostics} groups the diagnostic columns of Table~\ref{tab:residual_diagnostics} into a bar chart per method. The residual-active bars sit at 0.913 for Q and 0.888 for \method, the hard-condition bars at 2.959 and 2.534, and the HJB value, HJB advantage, and shield-active bars are nonzero only for \method, at 2.642, $-0.326$, and 0.120.


\begin{table}[htbp]
\centering
\caption{Coarse $20\times50$ short-horizon sweep ($N=1000$ episodes per method, 8~s each). Lower is better. Ref./path and altitude columns are in metres, airspeed in m/s, and Viol. frac. (safety-violation fraction) and Ctrl. idx. (runtime control-activity index) are dimensionless.}
\label{tab:coarse_sweep}
\footnotesize
\setlength{\tabcolsep}{3pt}
\begin{adjustbox}{max width=\textwidth}
\begin{tabular}{lrrrrrr}
\toprule
Method & Episodes & Ref./path RMS & Alt. RMS & Airspeed RMS & Ctrl. idx. & Viol. frac.\\
\midrule
Baseline & 1000 & 0.866 & 221.417 & 18.944 & 0.732 & 0.002223\\
Q residual & 1000 & 1.183 & 156.781 & 35.169 & 0.833 & 0.000447\\
HJB residual & 1000 & 1.174 & 155.149 & 35.303 & 0.822 & 0.000378\\
\bottomrule
\end{tabular}
\end{adjustbox}
\end{table}

Table~\ref{tab:coarse_sweep} reports the coarse $20\times50$ short-horizon sweep, averaging 1000 eight-second episodes per method. Here the ranking on mean spatial reference/path RMS inverts: the baseline is lowest at 0.866~m, \method\ is 1.174~m, and Q is 1.183~m. In the same sweep, mean altitude RMS falls from 221.417~m under the baseline to 156.781~m under Q and 155.149~m under \method, and mean airspeed RMS rises from 18.944~m/s under the baseline to 35.169~m/s under Q and 35.303~m/s under \method.

\begin{figure}[htbp]
\centering
\includegraphics[width=0.85\textwidth]{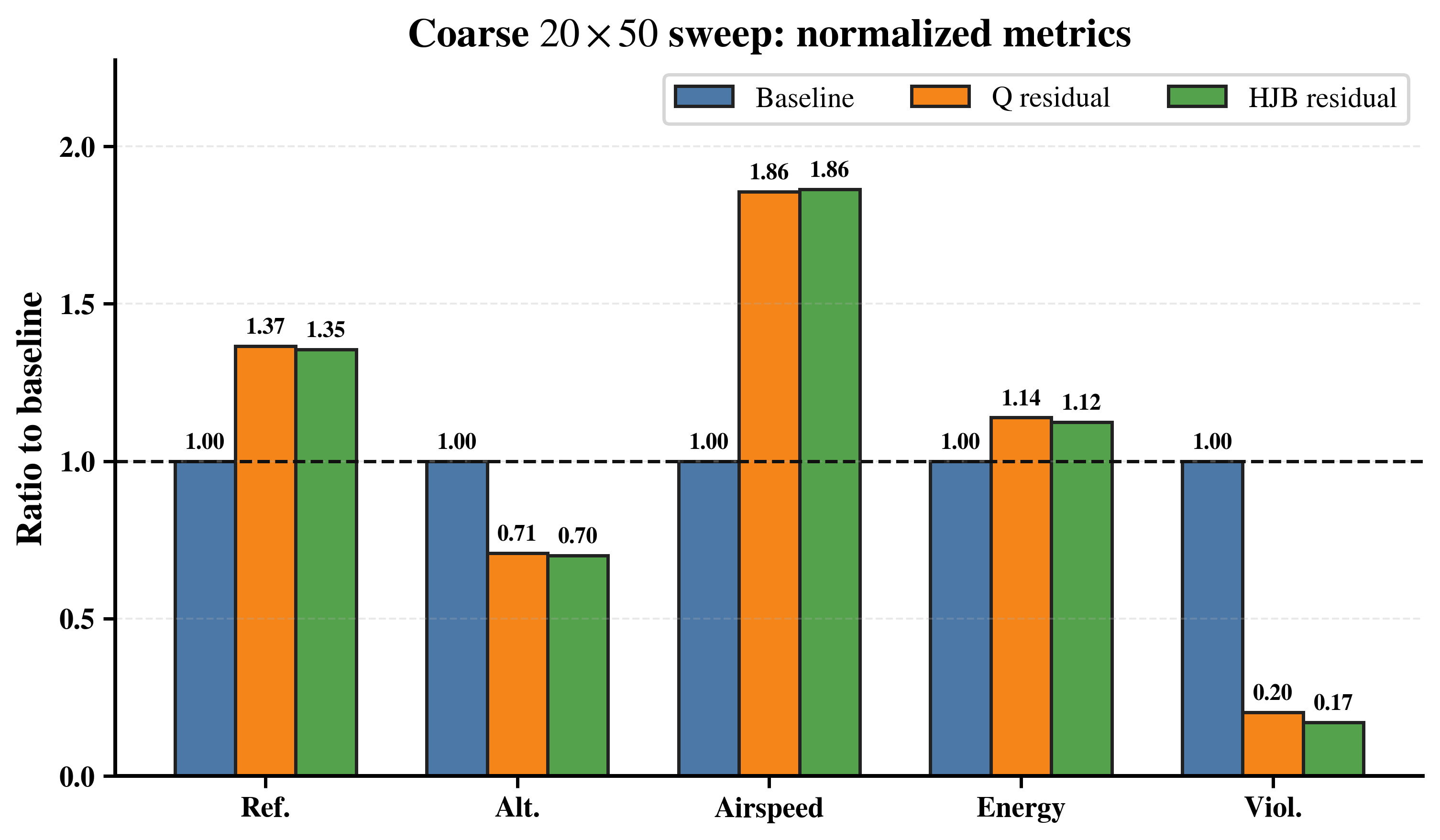}
\caption{Coarse $20\times50$ short-horizon sweep ($N=1000$ episodes per method, 8~s each). Lower is better. On mean spatial reference/path RMS the baseline is lowest (0.866~m), ahead of \method\ (1.174~m) and Q residual (1.183~m); altitude RMS falls under residual supervision and airspeed RMS rises.}
\label{fig:coarse_sweep}
\end{figure}
Figure~\ref{fig:coarse_sweep} plots the coarse-sweep means of Table~\ref{tab:coarse_sweep} as bars per metric. On spatial reference/path RMS the baseline bar is shortest at 0.866~m, with \method\ at 1.174~m and Q at 1.183~m, while the altitude bars fall from 221.417~m to 155.149~m under \method\ and the airspeed bars rise from 18.944~m/s to 35.303~m/s.


\begin{table}[htbp]
\centering
\caption{Fight-mode 60~s smoke run ($N=1$ run per method). Lower is better. Ref./path and altitude columns are in metres, airspeed in m/s, and Viol. frac. (safety-violation fraction), Ctrl. idx. (runtime control-activity index), and residual-active fraction are dimensionless.}
\label{tab:fight_smoke}
\footnotesize
\begin{adjustbox}{max width=\textwidth}
\begin{tabular}{lrrrrrr}
\toprule
Method & Ref./path RMS & Alt. RMS & Airspeed RMS & Ctrl. idx. & Viol. frac. & Residual active frac.\\
\midrule
Baseline & 177.140 & 100.880 & 8.703 & 13.532 & 0.003500 & 0.000\\
Q residual & 185.854 & 77.780 & 11.156 & 14.244 & 0.005000 & 0.078\\
HJB residual & 189.378 & 68.204 & 14.742 & 15.109 & 0.003333 & 0.194\\
\bottomrule
\end{tabular}
\end{adjustbox}
\end{table}

Table~\ref{tab:fight_smoke} reports the single 60~s fight-mode smoke run per method. The baseline has the lowest spatial reference/path RMS at 177.140~m, followed by Q at 185.854~m and \method\ at 189.378~m. \method\ has the lowest altitude RMS at 68.204~m, while its airspeed RMS is 14.742~m/s and its control-activity index is 15.109, both above the baseline values of 8.703~m/s and 13.532.

\begin{figure}[htbp]
\centering
\includegraphics[width=0.85\textwidth]{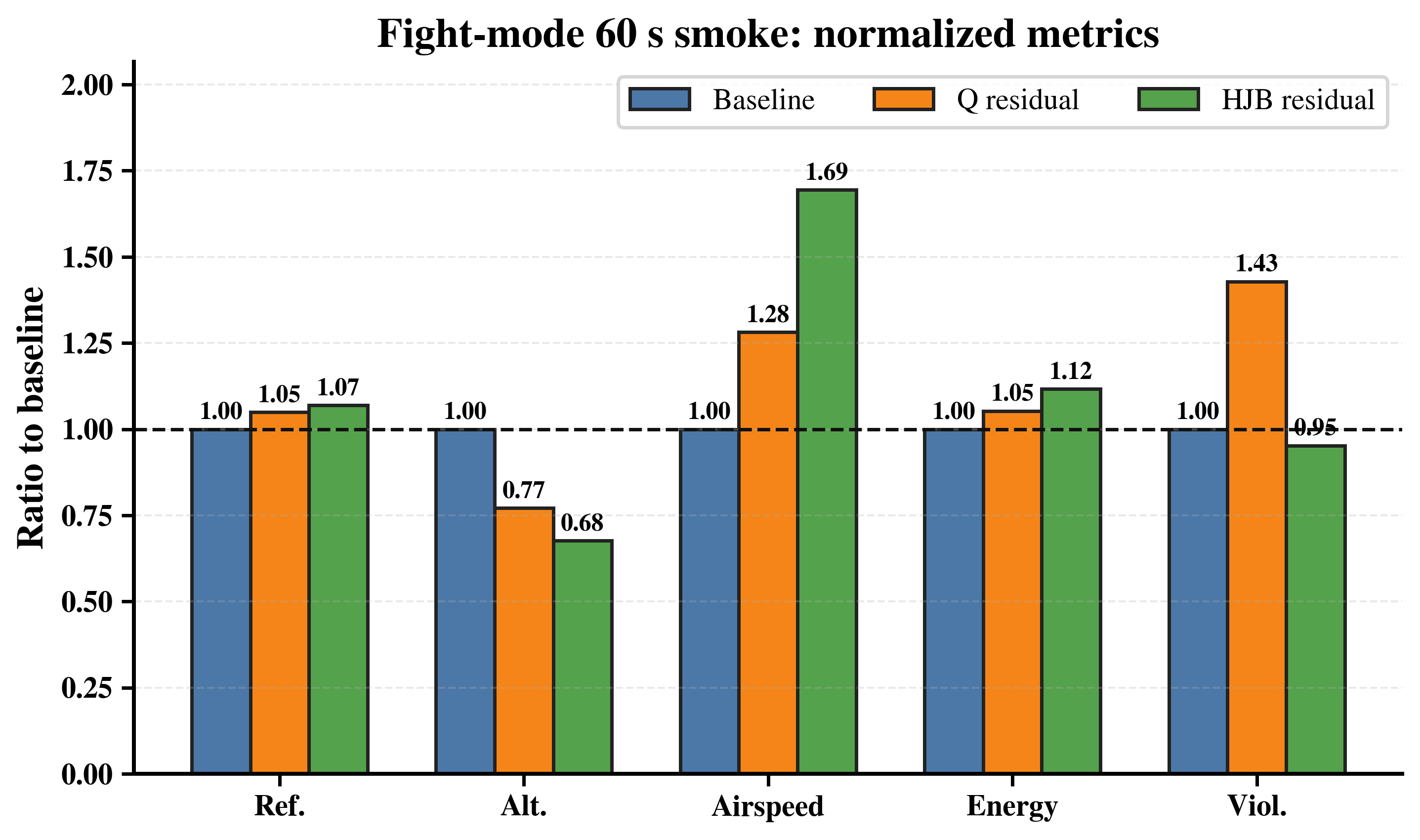}
\caption{Fight-mode 60~s smoke run ($N=1$ run per method). Lower is better. The baseline has the lowest spatial reference/path RMS (177.140~m), followed by Q residual (185.854~m) and \method\ (189.378~m); \method\ has the lowest altitude RMS.}
\label{fig:fight_smoke}
\end{figure}
Figure~\ref{fig:fight_smoke} plots the fight-mode smoke metrics of Table~\ref{tab:fight_smoke} as bars per method. The spatial reference/path bars rank the baseline lowest at 177.140~m, then Q at 185.854~m and \method\ at 189.378~m, while the altitude bars run the other way, with \method\ lowest at 68.204~m.

\begin{figure}[htbp]
\centering
\includegraphics[width=0.92\textwidth]{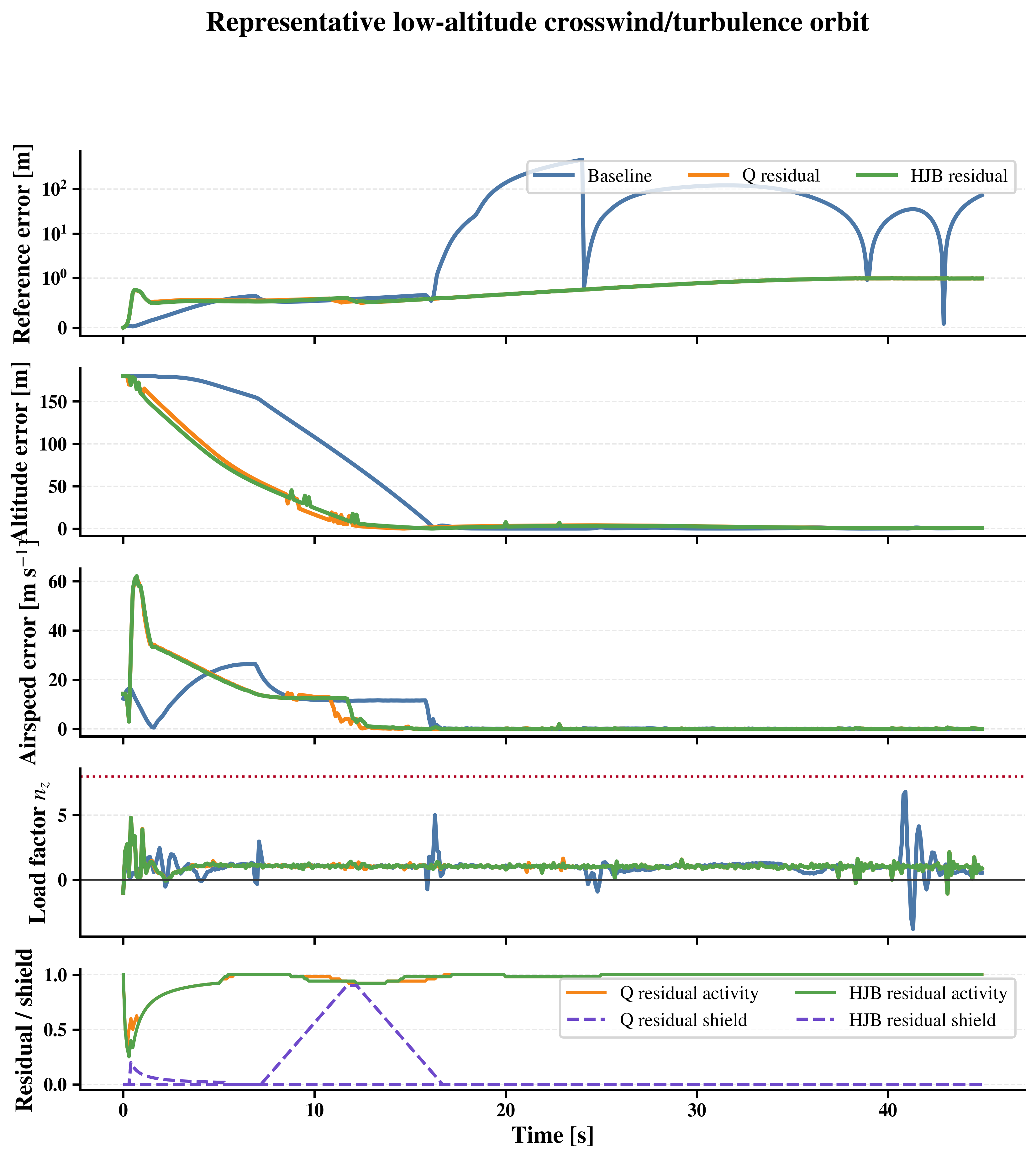}
\caption{Representative time series for one low-altitude crosswind/turbulence orbit scenario, showing reference error, altitude error, airspeed error, load factor, residual activity, and shield activity against time for the three methods. This is a single trace, not an aggregate.}
\label{fig:timeseries}
\end{figure}
\FloatBarrier
Figure~\ref{fig:timeseries} traces one low-altitude crosswind-and-turbulence orbit scenario over time, with reference error, altitude error, airspeed error, load factor, residual activity, and shield activity drawn for each method. The baseline reference-error trace stays large for the whole run, the two residual traces collapse toward zero once residual activity engages, and the residual and shield channels are nonzero only for the residual packages.

\section{Discussion}
\label{sec:discussion}

The benefit of residual command supervision is concentrated in long-horizon, high-disturbance flight, and it is small or absent elsewhere. Over the full-duration benchmark, \method\ lowers mean RMS reference error to 44.809~m against 338.617~m for the baseline and 88.809~m for the Q residual, an 86.77\% reduction relative to the baseline and a 49.54\% reduction relative to Q. The strength of this framing is that the gain comes from reshaping a reference rather than rewriting a controller: the airframe is still flown by the same inner loop, so the learner inherits the autopilot's stabilization, rate handling, and saturation behavior instead of having to relearn them. The corresponding weakness is that the aggregate gain is not a broad per-scenario win. The median reference error is a near-tie between Q and \method\ at 0.802~m, and most of the \method-over-Q mean gap traces to two adversarial fight scenarios in which the Q residual diverges while \method\ recovers: crosswind-with-turbulence and tailwind-with-lateral-gust, where Q reaches several hundred metres of reference error and the value-guided variant recovers the closed-loop trajectory. In the coarse early-phase sweep and the fight-mode smoke run the ranking inverts and the baseline tracks the reference best, with airspeed rising from 10.5 to 15.2~m/s once the residual engages. The picture the numbers support is narrow: when the mission is long, the disturbance is strong, and the baseline autopilot would otherwise drift far from its reference, a bounded value-guided residual reduces the closed-loop reference error; when the horizon is short or the air is nominal, the residual adds little and can track slightly worse.

That asymmetry has a mechanism, and the same mechanism explains both the gain and its cost. The residual never touches elevator, aileron, rudder, or throttle; it biases the airspeed, altitude, and heading reference the classical autopilot already tracks, toward energy recovery, altitude correction, or lateral recovery. Decoupling adaptive error reduction from low-level stabilization is the central reason small command-level corrections can move a path under disturbance: the learner targets the reference while the robust base controller absorbs the fast dynamics, an arrangement that constrained residual reinforcement learning has used to refine a nominal controller without taking on actuator-scale exploration risk, and that aircraft disturbance-rejection work motivates whenever compound aerodynamic effects accumulate against a fixed trajectory. When that bias is well aimed in a hard scenario, spatial and altitude error fall because the autopilot is no longer asked to hold only the nominal mission command. Speed, altitude, lateral motion, throttle, pitch, and load are coupled, so the same reference bias spends pitch and throttle authority that would otherwise regulate airspeed. The residual modes therefore trade reduced path and altitude error for increased airspeed error and higher control activity: airspeed RMS rises from 10.546~m/s at the baseline to 15.086~m/s under Q and 15.191~m/s under \method, and the control-activity index rises with it. This is a physical trade-off rather than a tuning artifact. Aggressive flight-path and altitude correction couples directly into thrust use, so a controller that prioritizes path tends to saturate the energy channel that would otherwise hold speed, which is why classical designs reach for total-energy management and anti-windup structure to keep the two channels balanced. The cost is visible directly in the per-scenario winner counts, where the baseline wins every scenario on airspeed.

The advantage of the value layer over a purely tabular residual is that it scores each candidate against doing nothing, and that scoring is what separates \method\ from Q in the hardest cases. A raw Q estimate ranks actions by absolute value and can mis-rank when the useful signal is small relative to the table's noise; anchoring the decision to a no-op-relative advantage makes the comparison sharper, in the spirit of advantage-based action selection. The diagnostics show where this engages without certifying why it helps. The residual-active fraction near 0.9 shows the supervisor selects a nonzero residual on most steps once the hard-condition gate opens; the shield-active fraction of about 0.12 under \method\ shows the finite-action shield removing candidates on roughly one step in eight. The mean Hamiltonian advantage of \(-0.326\) is a no-op-relative score: selected residuals scored better than preserving the nominal command under the value proxy, on average. These quantities confirm that the value and filtering machinery was exercised; they are surrogate signals over a hand-designed predictor, not evidence of forward invariance or optimality. Tabular residual supervision over discretized tracking, wind, and energy features is itself a reasonable design, with finite-state abstraction and Q-learning attitude control on agile aircraft as precedent \citep{taherian2014,zahmatkesh2022}, but the benchmark shows its ceiling: the Q residual without the shield is the variant that diverges in the two fight scenarios that the value-guided selector survives.

The baseline remains a strong reference precisely on the channels the residual sacrifices, and it should be read that way. The small-UAV modeling and control basis it follows keeps nested actuator-facing loops with bandwidth separation and saturation handling intact \citep{beard_mclain_ngc}, and gain-scheduled fixed-wing autopilots carry that structure across flight conditions while staying re-tunable \citep{poksawat2017}, an approach that also extends to airframes with minimal control surfaces \citep{liu2015}. Because the baseline does not chase residual geometric corrections, it is less aggressive on difficult path cases but protects airspeed regulation and uses less control activity. The same wind and turbulence literature that drives our hard scenarios reaches the trade-offs we observe by altering the low-level controller instead of the reference: steady-wind trajectory design shows that wind reshapes which maneuvers are feasible at all \citep{ayhan2018}, low-altitude wind shear degrades curved-path following enough to justify dedicated guidance \citep{zhang2024}, and turbulence-mitigation work redesigns the inner loop through segmented surfaces or a robust nonlinear law \citep{sattar2022,meharie2024}. Our contribution is to ask whether bounded reshaping of the reference alone can recover difficult tracking cases while that inner loop stays fixed, and the answer is a qualified yes confined to the long-horizon regime.

Against actuator-level residual RL, \method\ is deliberately more conservative. The canonical formulation injects the learned correction at the torque or actuator interface \citep{johannink2019}, and aerospace residual learning carries that idea into flight while arguing for explicit safety machinery around the learned term \citep{jayarathne2023}; supervisory RL in another safety-critical plant places the learned decision above the plant-facing loop rather than inside it \citep{sun2024}, which is the posture we adopt. By correcting only the reference, each intervention is bounded by command projection, inspectable at every step, and disengageable through the no-op action, at the cost of authority limited by the autopilot and actuator response. This command-layer view is shared with control-theoretic methods that treat the reference itself as the design surface, through command-filtered backstepping \citep{dong2011}, modified command and reference models \citep{na2019}, and command limiting \citep{sun2021}; what we add is a finite residual on commanded airspeed, altitude, and heading with a shared-runtime comparison against both a classical autopilot and a Q-only residual, reporting the airspeed and control-activity cost the intervention incurs. The broader case for keeping a robust base controller and layering learning above it, rather than replacing it, is also made by work that integrates adaptive control with reinforcement learning \citep{annaswamy2023} and that hardens learned policies with an adaptive inner controller \citep{cheng2022}.

The relationship to safe RL, control barrier functions, and HJB or ADP methods is one of inspiration, not equivalence. The admissible-action restriction follows constraint-admissible safe RL \citep{li2018}. The value-growth and risk gates borrow the pointwise stability-and-safety conditions that control-barrier-function quadratic programs enforce \citep{ames2016}, with the approximate optimal and scalable variants of that filtering as further precedent \citep{cohen2020,gurriet2020}, and the barrier-Lyapunov actor-critic showing how such conditions enter a learned critic \citep{zhao2023}. The value-based safety reasoning is motivated by Hamilton-Jacobi reachability bridged with RL \citep{fisac2019}, whose exactness scales poorly with state dimension. The Hamiltonian advantage and the semi-discrete value iteration draw on Hamiltonian-driven approximate dynamic programming \citep{yang2017}, including its treatment of approximation error \citep{yang2021} and its experience-replay extension \citep{yang2022}, and on HJB optimal feedback with explicit action constraints \citep{lutter2020}. The notion of a learned critic acting as a hard constraint comes from safe-value functions \citep{tan2024}. In all of these our construction operates as a finite-action shield over predicted features rather than a solved continuous HJB partial differential equation or a certified barrier; the negative mean advantage and the nonzero shield fraction are surrogate diagnostics, and the tie-heavy, nonuniform safety-violation fraction across scenarios discourages any overbroad safety reading of the aggregate result.

These results are bounded in ways that shape how far they can be carried. The study is entirely in simulation, with no real-flight, hardware-in-the-loop, software-timing, or wind-tunnel evidence, and the aircraft is a coefficient-based small-UAV model rather than a calibrated airframe, its nominal trim speed and motor constant set as repository stress-test values rather than certified limits; simulation environments are known to yield more optimistic robustness than physical flight, which is why we treat the digital twin as an engineering reference and not a high-fidelity aerodynamic surrogate. There is no formal stability or safety certificate: the value-iteration critic and the CLF- and CBF-inspired finite-action shield are heuristic constructions over normalized features, not a continuous HJB solution or a certified forward-invariant barrier, so command boundedness from projection and nominal baseline invariance under the hard-condition gate are the only structural properties we assert. The comparison is at the level of the controller package rather than a component ablation, because a deterministic energy-allocation helper is active in both residual modes but not in the baseline, so the residual-versus-baseline gain cannot be attributed to the value or critic layer alone, and the individual contributions of the Q table, the value proxy, the Hamiltonian advantage, the finite-action shield, the hard-condition gate, the action set, the reward shaping, and the energy helper are not separately isolated. The primary full-duration benchmark uses a single seed per scenario, so its across-scenario intervals are descriptive rather than repeated-seed confidence intervals, and the only multi-run evidence is the short early-phase sweep and a single fight-mode run; the seed-sensitivity literature is explicit that this is supplementary robustness evidence, not full-mission validation \citep{eimer2023}. Within these bounds the benchmark is useful because it is code-traceable and reports the cost of residual supervision alongside its benefit, which is what makes the long-horizon, high-disturbance gain interpretable rather than incidental, and it points at the obvious next steps: per-component ablation of the helper, shield, advantage, and reward terms, full-duration multi-seed trials, and a path toward hardware-in-the-loop and flight evidence.


\section{Conclusion}
\label{sec:conclusion}
We presented \method, an autopilot-preserving command supervisor that leaves a classical fixed-wing autopilot as the actuator-facing controller and confines learning to bounded residuals on the commanded airspeed, altitude, and heading. The residual is drawn from a finite action set and scored by a tabular Q-value, a semi-discrete HJB/Bellman value proxy, a no-op-relative Hamiltonian advantage, and a CLF/CBF-inspired finite-action shield, with the no-op action always retained as a fallback. The supervisor never reaches the elevator, aileron, rudder, or throttle; it only reshapes the reference the autopilot already tracks.

We evaluated the method in a shared simulation runtime in which the baseline autopilot, the tabular Q residual, and \method\ pass through the same nonlinear plant, actuator model, mission generator, and disturbance and metric definitions, so that the comparison is at the level of the whole controller package; the two residual supervisors also share a deterministic energy-allocation helper that the baseline lacks, so the baseline-versus-residual gain reflects the residual package as a whole rather than the residual-decision layer in isolation. The primary evidence is a full-duration 20-scenario benchmark; a short-horizon early-phase sweep and a fight-mode 60~s run provide supporting context for seed sensitivity and finite-runtime behavior rather than additional performance claims.

On the full-duration benchmark, \method\ attains a mean RMS reference error of 44.809~m, against 338.617~m for the baseline autopilot and 88.809~m for the tabular Q residual. This is an 86.77\% reduction relative to the classical baseline and a 49.54\% reduction relative to Q. The improvement is concentrated in spatial and altitude tracking under disturbance and does not extend to every metric: airspeed RMS error and control activity rise rather than fall, so the supervisor reshapes the trade-off rather than dominating it. Relative to the classical autopilot it preserves the inner loop while lowering long-horizon, high-disturbance reference error; the claim is therefore narrow, that bounded value-guided shielded supervision improves difficult reference tracking in this benchmark without replacing the underlying controller.

Several directions would strengthen and sharpen this evidence. Repeating the full-duration benchmark with multiple seeds per scenario would turn the reported intervals from descriptive into inferential. A component ablation that isolates the value-iteration critic and the finite-action shield from the deterministic energy-allocation helper active in the residual modes would attribute the observed gain to its sources. Beyond the present coefficient-based model, higher-fidelity dynamics and software- and hardware-in-the-loop studies would test the supervisor under more realistic actuation and sensing, and controlled flight trials would assess transfer to physical aircraft.

\section*{Acknowledgements}
This article was carried out within the scope of the T\"{U}B\.{I}TAK 3501 project numbered 225M067, titled ``A Next-Generation Flight Control Architecture for Fixed-Wing UAVs: Online Uncertainty and Parameter Adaptation with Hardware-in-the-Loop Integrated HJB-SAC Control.''

\section*{Data availability statement}
The tables and figures in this paper are derived from the simulation artifacts produced by the benchmark runs. The CSV, JSON, and JSONL files identified in the figure captions and appendices contain the per-method aggregates, per-scenario results, and diagnostic logs underlying every reported number.

\section*{Code availability statement}
The implementation is openly available in the GitHub repository at \url{https://github.com/PhiniteLab/pythalab-sharq-hjb-uav-command-supervision} \citep{phinitelab_sharq_hjb_repo}, with a companion demonstration video at \url{https://youtu.be/CIf4Dxo5INg}. The repository contains the nonlinear fixed-wing plant and atmosphere model, the classical autopilot and actuator model, the shared simulation runtime, the tabular residual Q-learning supervisor, the HJB-inspired value-iteration critic and Hamiltonian-advantage scoring, the CLF/CBF-style finite-action shield, the deterministic energy-allocation helper, the scenario and mission generators, and the experiment runner and metric definitions used to produce the reported artifacts. Readers may use the repository for testing, reproduction, and derivative research provided that the repository and this paper are cited.

\section*{Declaration on the use of generative AI tools}
The authors used a generative-AI assistant solely to improve the English-language quality of the manuscript, that is, to correct grammar and refine the phrasing and clarity of text written by the authors. It was not used to generate, design, or analyze any of the methods, code, experiments, data, or results. The underlying technical work was carried out by the authors and can be independently examined in the public GitHub repository and the companion demonstration video cited in the Code availability statement.

\section*{Conflict of interest statement}
The authors declare no conflict of interest.


\appendix
\section{Metric implementation trace}
\label{app:metric_implementation_trace}
All metrics are computed from the same telemetry definitions for every method.  The spatial reference/path RMS aggregates the position and reference-tracking error of the benchmark, the altitude and airspeed RMS terms report channel-specific tracking error, the control-activity index is an episode integral of the runtime control effort, and the maximum load factor is the largest absolute logged $n_z$.  The safety-violation fraction is the fraction of simulated time in which the runtime safety-condition check is active.

Episode summaries are written by the runtime execution path and then aggregated by method for the manuscript tables.  The diagnostic metrics, including residual-active fraction, shield-active fraction, HJB value, HJB advantage, HJB stage cost, hard-condition score, and candidate count, are runtime-reported quantities from the residual supervisors.  A zero under a method that lacks the corresponding layer marks the absence of that layer, not a measured physical zero.

\section{Software regression and artifact-generation checks}
\label{app:software_regression_checks}
The repository test suite documents the software conditions under which the artifacts were generated.  The core tests in \texttt{backend/tests/test\_uavsim\_core.py} exercise deterministic reproduction of the plant, atmosphere, and wind model, consistency of the actuator, telemetry, and command interfaces, and end-to-end benchmark execution with the expected summary outputs.  The server tests in \texttt{backend/tests/test\_uavsim\_server.py} check request handling, rejection of non-finite or invalid configurations, profile-reset state clearing, and stable streaming of simulation frames.

These tests are smoke and integration checks for traceability and regression control.  They confirm that the artifacts come from a finite, shared software pipeline with explicit clamps and schema checks.  Higher-fidelity aerodynamic validation, hardware-in-the-loop testing, flight testing, and formal verification are outside their scope.

\section{Scenario catalog}
\label{app:scenario_catalog}
The scenario catalog lists the geometry and the environmental stressors of each benchmark scenario so that the conditions are fully traceable. Every full-duration scenario is fixed by a mission-profile label, a random seed, a duration, commanded airspeed and altitude, geometry parameters such as orbit diameter where applicable, and a disturbance block of steady north, east, and down wind components, body-axis gusts, and turbulence intensity.

The same scenario definition is reused for the baseline autopilot, the tabular-Q residual supervisor, and \method. Because the conditions are matched, differences across the tables and figures reflect controller behavior under identical scenarios rather than different disturbances or episode lengths. The two stress mechanisms are listed separately. Mission geometry stresses lateral and longitudinal tracking through tight turns, racetrack segments, climb and descent transitions, low-altitude recovery, and fight-mode S-turns, while the environmental fields stress robustness through crosswind, headwind, tailwind, vertical gusts, diagonal gusts, and turbulence. The main text reports the aggregate results; the tables below give the scenario-level inputs behind them.

The per-scenario comparisons are paired comparisons of the three controller packages in a common runtime, not independent flights or separate aerodynamic models.

\begingroup
\scriptsize
\setlength{\tabcolsep}{3pt}
\renewcommand{\arraystretch}{0.98}
\begin{longtable}{r P{0.24\textwidth}P{0.20\textwidth}rrrrr}
\caption{Mission geometry and timing of the 20 benchmark scenarios: seed, duration (s), commanded altitude \(h_c\) (m), orbit diameter (m), and commanded airspeed \(V_{a,c}\) (m s\(^{-1}\)).}\label{tab:scenario_geometry}\\
\toprule
\# & Scenario & Profile & Seed & Dur. & $h_c$ & Diam. & $V_{a,c}$\\
\midrule
\endfirsthead
\toprule
\# & Scenario & Profile & Seed & Dur. & $h_c$ & Diam. & $V_{a,c}$\\
\midrule
\endhead
1 & orbit crosswind turbulence low alt & loiter orbit & 4101 & 45 & 180 & 200 & 30\\
2 & tight orbit tailwind gust & high speed climb s turn 200 & 4102 & 45 & 210 & 160 & 34\\
3 & wide orbit headwind energy & racetrack & 4103 & 50 & 240 & 320 & 45\\
4 & reverse orbit shear & figure eight & 4104 & 45 & 200 & 220 & 32\\
5 & high altitude headwind climb & straight climb altitude hold & 4105 & 55 & 300 & 260 & 40\\
6 & low alt vertical gust recovery & takeoff climbout 200 & 4106 & 45 & 140 & 180 & 28\\
7 & fast circle mixed crosswind & runway takeoff accel 200 & 4107 & 50 & 250 & 300 & 55\\
8 & small radius crosswind orbit & loiter orbit & 4108 & 45 & 200 & 140 & 32\\
9 & fight vertical gust & fight mode & 4109 & 55 & 240 & 480 & 112\\
10 & fight crosswind turbulence & fight mode & 4110 & 55 & 240 & 480 & 112\\
11 & oblique takeoff crosswind & runway takeoff accel 200 & 4111 & 45 & 170 & 240 & 34\\
12 & steep climb headwind turbulence & takeoff climbout 200 & 4112 & 60 & 340 & 280 & 42\\
13 & long straight crosswind rejoin & straight climb altitude hold & 4113 & 55 & 220 & 260 & 38\\
14 & reverse fast orbit tailwind & racetrack & 4114 & 50 & 230 & 260 & 52\\
15 & low speed vertical shear hold & loiter orbit & 4115 & 45 & 190 & 220 & 24\\
16 & compact circle windshift & figure eight & 4116 & 45 & 210 & 150 & 30\\
17 & high speed climb crosswind & high speed climb s turn 200 & 4117 & 55 & 280 & 300 & 58\\
18 & fight tailwind lateral gust & fight mode & 4118 & 55 & 240 & 480 & 112\\
19 & fight diagonal max turbulence & fight mode & 4119 & 55 & 240 & 480 & 112\\
20 & racetrack reverse vertical mixed & racetrack & 4120 & 50 & 260 & 240 & 44\\
\bottomrule
\end{longtable}
\endgroup

\begingroup
\scriptsize
\setlength{\tabcolsep}{3pt}
\renewcommand{\arraystretch}{0.98}
\begin{longtable}{r P{0.22\textwidth}rrrrrr P{0.30\textwidth}}
\caption{Environmental stressors of the 20 benchmark scenarios: steady wind components \(w_N,w_E,w_D\) (m s\(^{-1}\)) and body-axis gust components \(g_u,g_v,g_w\) (m s\(^{-1}\)), with a short scenario description.}\label{tab:scenario_disturbance}\\
\toprule
\# & Scenario & $w_N$ & $w_E$ & $w_D$ & $g_u$ & $g_v$ & $g_w$ & Description\\
\midrule
\endfirsthead
\toprule
\# & Scenario & $w_N$ & $w_E$ & $w_D$ & $g_u$ & $g_v$ & $g_w$ & Description\\
\midrule
\endhead
1 & orbit crosswind turbulence low alt & 0.0 & 10.0 & 0.0 & 0.0 & 3.0 & 0.0 & Low-altitude 200 m orbit with maximum crosswind and turbulence.\\
2 & tight orbit tailwind gust & 10.0 & 0.0 & 0.0 & 0.0 & 4.0 & 0.0 & Tighter orbit and tailwind/gust stress after climb-out.\\
3 & wide orbit headwind energy & -10.0 & 0.0 & 0.0 & -3.5 & 0.0 & 0.0 & Wide faster orbit under strong headwind to stress total-energy recovery.\\
4 & reverse orbit shear & 0.0 & -10.0 & 3.0 & 0.0 & -4.0 & 0.0 & Reverse-turn orbit with vertical wind component and lateral shear proxy.\\
5 & high altitude headwind climb & -8.0 & 0.0 & 0.0 & -4.0 & 0.0 & 2.5 & Higher target altitude and headwind during climb/straight/orbit transition.\\
6 & low alt vertical gust recovery & 0.0 & 0.0 & 0.0 & -2.0 & 0.0 & 4.0 & Lower-altitude reference with maximum vertical gust and turbulence.\\
7 & fast circle mixed crosswind & 8.0 & -8.0 & 0.0 & 3.0 & -3.0 & 0.0 & Fast 300 m circle with diagonal wind and gust coupling.\\
8 & small radius crosswind orbit & 0.0 & 10.0 & 0.0 & 0.0 & -4.0 & 0.0 & Small-radius orbit under lateral gusts, intentionally lateral-error dominated.\\
9 & fight vertical gust & -9.0 & 0.0 & 0.0 & 0.0 & 0.0 & 4.0 & Fight-mode S-turn with vertical gusts and headwind.\\
10 & fight crosswind turbulence & 0.0 & -10.0 & 0.0 & -3.0 & 4.0 & 0.0 & Fight-mode S-turn with maximum crosswind/turbulence and mixed gusts.\\
11 & oblique takeoff crosswind & 6.0 & 9.0 & 0.0 & 0.0 & 0.0 & 2.5 & Runway takeoff with diagonal crosswind and mild vertical gust before orbit capture.\\
12 & steep climb headwind turbulence & -10.0 & 0.0 & 0.0 & -4.0 & 0.0 & 3.0 & Higher climb target under headwind/turbulence to stress energy allocation.\\
13 & long straight crosswind rejoin & 0.0 & 10.0 & 0.0 & 0.0 & 3.5 & 0.0 & Long straight leg before orbit entry with sustained crosswind rejoin demand.\\
14 & reverse fast orbit tailwind & 9.0 & 0.0 & 0.0 & 0.0 & -4.0 & 0.0 & Reverse fast orbit with tailwind and lateral gust forcing.\\
15 & low speed vertical shear hold & 0.0 & 0.0 & -4.0 & 0.0 & 0.0 & 4.0 & Lower-speed orbit with vertical shear proxy and high turbulence.\\
16 & compact circle windshift & -7.0 & 7.0 & 0.0 & 4.0 & -4.0 & 0.0 & Compact circle under mixed north/east wind and opposing body gusts.\\
17 & high speed climb crosswind & 0.0 & -10.0 & 0.0 & 0.0 & 3.0 & -3.5 & Fast climb/orbit entry with strong crosswind and positive vertical gust.\\
18 & fight tailwind lateral gust & 10.0 & 0.0 & 0.0 & 0.0 & -4.0 & 2.5 & Fight-mode S-turn with tailwind and alternating lateral gust stress.\\
19 & fight diagonal max turbulence & -8.0 & 8.0 & 0.0 & -4.0 & 4.0 & 4.0 & Fight-mode S-turn with diagonal wind, mixed gusts, and maximum turbulence.\\
20 & racetrack reverse vertical mixed & -6.0 & -6.0 & 4.0 & 3.0 & 0.0 & -3.0 & Reverse-turn racetrack-like orbit with vertical wind and mixed gusts.\\
\bottomrule
\end{longtable}
\endgroup


\section{Per-scenario full-duration results}
\label{app:per_scenario_full_duration_results}
The tables below give the scenario-level tracking and operational metrics behind the full-duration aggregate and are read together with the Results section.

\begingroup
\scriptsize
\setlength{\tabcolsep}{2.5pt}
\renewcommand{\arraystretch}{0.98}
\begin{longtable}{r P{0.21\textwidth}rrrrrrrrr}
\caption{Per-scenario tracking metrics over the full-duration benchmark ($N=20$ scenarios, one episode each). Ref./path and altitude columns are in metres, airspeed columns in m/s. Lower is better. B = baseline, Q = tabular Q residual, SH = \method.}\label{tab:per_scenario_tracking}\\
\toprule
\# & Scenario & Ref./path B & Ref./path Q & Ref./path SH & Alt. B & Alt. Q & Alt. SH & Air B & Air Q & Air SH\\
\midrule
\endfirsthead
\toprule
\# & Scenario & Ref./path B & Ref./path Q & Ref./path SH & Alt. B & Alt. Q & Alt. SH & Air B & Air Q & Air SH\\
\midrule
\endhead
1 & orbit crosswind turbulence low alt & 108.88 & 0.77 & 0.77 & 79.89 & 49.34 & 47.07 & 9.37 & 12.39 & 12.47\\
2 & tight orbit tailwind gust & 124.15 & 151.67 & 149.95 & 96.32 & 57.19 & 56.84 & 8.98 & 14.74 & 14.73\\
3 & wide orbit headwind energy & 640.77 & 0.74 & 0.67 & 107.30 & 69.27 & 66.83 & 9.41 & 15.11 & 15.15\\
4 & reverse orbit shear & 98.94 & 0.79 & 0.79 & 99.76 & 59.56 & 57.23 & 10.84 & 14.28 & 14.30\\
5 & high altitude headwind climb & 643.54 & 0.65 & 0.65 & 156.88 & 91.46 & 88.91 & 11.44 & 18.02 & 18.13\\
6 & low alt vertical gust recovery & 92.20 & 0.71 & 0.69 & 79.02 & 47.90 & 49.19 & 12.30 & 13.96 & 15.02\\
7 & fast circle mixed crosswind & 538.10 & 0.86 & 0.86 & 114.03 & 67.04 & 66.82 & 8.87 & 15.38 & 15.32\\
8 & small radius crosswind orbit & 142.68 & 0.77 & 0.76 & 124.08 & 62.23 & 59.95 & 12.74 & 14.97 & 14.99\\
9 & fight vertical gust & 1213.68 & 0.67 & 0.67 & 130.44 & 68.61 & 68.58 & 12.09 & 15.43 & 15.42\\
10 & fight crosswind turbulence & 420.69 & 493.08 & 57.41 & 140.15 & 69.60 & 69.12 & 11.99 & 15.80 & 15.74\\
11 & oblique takeoff crosswind & 172.95 & 0.80 & 0.80 & 93.27 & 49.24 & 48.98 & 11.75 & 13.30 & 13.35\\
12 & steep climb headwind turbulence & 330.15 & 0.66 & 0.78 & 177.61 & 102.84 & 100.24 & 11.46 & 19.22 & 19.40\\
13 & long straight crosswind rejoin & 84.58 & 157.47 & 156.19 & 93.00 & 55.38 & 55.03 & 8.89 & 13.30 & 13.36\\
14 & reverse fast orbit tailwind & 155.03 & 211.07 & 191.74 & 101.69 & 60.64 & 60.57 & 8.61 & 14.60 & 14.46\\
15 & low speed vertical shear hold & 108.65 & 132.25 & 115.29 & 86.23 & 53.81 & 54.32 & 9.31 & 13.28 & 14.08\\
16 & compact circle windshift & 168.55 & 0.77 & 0.75 & 117.72 & 64.35 & 61.79 & 12.25 & 14.99 & 15.11\\
17 & high speed climb crosswind & 545.21 & 0.82 & 0.82 & 126.16 & 79.37 & 76.93 & 9.06 & 16.36 & 16.35\\
18 & fight tailwind lateral gust & 559.20 & 620.02 & 214.96 & 114.00 & 67.27 & 64.77 & 9.82 & 15.15 & 14.99\\
19 & fight diagonal max turbulence & 536.10 & 0.81 & 0.81 & 104.33 & 65.16 & 65.02 & 9.98 & 14.71 & 14.72\\
20 & racetrack reverse vertical mixed & 88.31 & 0.80 & 0.80 & 140.86 & 76.11 & 76.03 & 11.72 & 16.72 & 16.71\\
\bottomrule
\end{longtable}
\endgroup

\begingroup
\scriptsize
\setlength{\tabcolsep}{2.5pt}
\renewcommand{\arraystretch}{0.98}
\begin{longtable}{r P{0.21\textwidth}rrrrrrrrr}
\caption{Per-scenario operational metrics over the full-duration benchmark ($N=20$ scenarios, one episode each). Lower is better. B = baseline, Q = tabular Q residual, SH = \method, Viol. = safety-violation fraction (dimensionless), Ctrl. = runtime control-activity index, and $|n_z|$ = maximum absolute load factor.}\label{tab:per_scenario_operational}\\
\toprule
\# & Scenario & Ctrl. B & Ctrl. Q & Ctrl. SH & Viol. B & Viol. Q & Viol. SH & $|n_z|$ B & $|n_z|$ Q & $|n_z|$ SH\\
\midrule
\endfirsthead
\toprule
\# & Scenario & Ctrl. B & Ctrl. Q & Ctrl. SH & Viol. B & Viol. Q & Viol. SH & $|n_z|$ B & $|n_z|$ Q & $|n_z|$ SH\\
\midrule
\endhead
1 & orbit crosswind turbulence low alt & 4.24 & 5.30 & 5.30 & 0.0040 & 0.0000 & 0.0000 & 8.00 & 6.53 & 6.51\\
2 & tight orbit tailwind gust & 3.38 & 3.51 & 3.43 & 0.0011 & 0.0000 & 0.0000 & 8.00 & 6.59 & 6.31\\
3 & wide orbit headwind energy & 4.76 & 6.42 & 6.29 & 0.0000 & 0.0000 & 0.0000 & 5.85 & 5.61 & 5.52\\
4 & reverse orbit shear & 3.70 & 5.42 & 5.39 & 0.0011 & 0.0000 & 0.0000 & 8.00 & 6.68 & 6.87\\
5 & high altitude headwind climb & 5.61 & 7.55 & 7.50 & 0.0004 & 0.0000 & 0.0000 & 6.62 & 6.69 & 6.35\\
6 & low alt vertical gust recovery & 2.90 & 5.51 & 5.30 & 0.0004 & 0.0013 & 0.0013 & 6.64 & 7.54 & 7.52\\
7 & fast circle mixed crosswind & 4.45 & 6.26 & 6.24 & 0.0024 & 0.0000 & 0.0000 & 8.00 & 6.61 & 6.61\\
8 & small radius crosswind orbit & 3.71 & 5.65 & 5.54 & 0.0020 & 0.0011 & 0.0002 & 7.40 & 7.42 & 6.81\\
9 & fight vertical gust & 12.74 & 7.36 & 7.36 & 0.0435 & 0.0000 & 0.0000 & 8.00 & 5.93 & 5.84\\
10 & fight crosswind turbulence & 11.19 & 11.72 & 8.95 & 0.0040 & 0.0189 & 0.0153 & 8.00 & 8.00 & 8.00\\
11 & oblique takeoff crosswind & 3.92 & 5.37 & 5.36 & 0.0020 & 0.0000 & 0.0000 & 8.00 & 8.00 & 8.00\\
12 & steep climb headwind turbulence & 4.96 & 8.74 & 8.78 & 0.0018 & 0.0000 & 0.0000 & 8.00 & 8.00 & 8.00\\
13 & long straight crosswind rejoin & 4.38 & 4.28 & 4.41 & 0.0035 & 0.0000 & 0.0000 & 8.00 & 7.19 & 7.19\\
14 & reverse fast orbit tailwind & 3.85 & 4.25 & 4.18 & 0.0010 & 0.0000 & 0.0000 & 8.00 & 5.40 & 6.09\\
15 & low speed vertical shear hold & 2.53 & 2.72 & 2.62 & 0.0004 & 0.0000 & 0.0000 & 6.39 & 7.07 & 7.02\\
16 & compact circle windshift & 3.62 & 5.56 & 5.53 & 0.0000 & 0.0004 & 0.0000 & 5.92 & 6.27 & 6.13\\
17 & high speed climb crosswind & 4.74 & 7.42 & 7.37 & 0.0018 & 0.0000 & 0.0000 & 7.21 & 7.41 & 7.41\\
18 & fight tailwind lateral gust & 11.83 & 13.60 & 13.51 & 0.0051 & 0.0400 & 0.0498 & 8.00 & 8.00 & 8.00\\
19 & fight diagonal max turbulence & 12.72 & 7.29 & 7.29 & 0.0209 & 0.0000 & 0.0000 & 8.00 & 6.64 & 6.64\\
20 & racetrack reverse vertical mixed & 4.04 & 6.55 & 6.55 & 0.0006 & 0.0000 & 0.0000 & 6.25 & 5.30 & 5.19\\
\bottomrule
\end{longtable}
\endgroup


\section{Coarse early-phase sweep and fight-mode smoke evidence}
\label{app:coarse_and_smoke_evidence}
The coarse \(20\times 50\) sweep and the 60 s fight-mode run are repeated here for completeness. The sweep reports short-horizon early-phase repetition behavior, and the fight-mode run is a finite-runtime smoke check.


\begin{table}[htbp]
\centering
\caption{Supplementary coarse $20\times50$ short-horizon sweep ($N=1000$ episodes per method, 8~s each). Lower is better. Ref./path and altitude columns are in metres, airspeed in m/s, and Viol. frac. (safety-violation fraction) and Ctrl. idx. (runtime control-activity index) are dimensionless.}
\label{tab:coarse_sweep_appendix}
\footnotesize
\setlength{\tabcolsep}{3pt}
\begin{adjustbox}{max width=\textwidth}
\begin{tabular}{lrrrrrr}
\toprule
Method & Episodes & Ref./path RMS & Alt. RMS & Airspeed RMS & Ctrl. idx. & Viol. frac.\\
\midrule
Baseline & 1000 & 0.866 & 221.417 & 18.944 & 0.732 & 0.002223\\
Q residual & 1000 & 1.183 & 156.781 & 35.169 & 0.833 & 0.000447\\
HJB residual & 1000 & 1.174 & 155.149 & 35.303 & 0.822 & 0.000378\\
\bottomrule
\end{tabular}
\end{adjustbox}
\end{table}


\begin{table}[htbp]
\centering
\caption{Supplementary fight-mode 60~s smoke run ($N=1$ run per method). Lower is better. Ref./path and altitude columns are in metres, airspeed in m/s, and Viol. frac. (safety-violation fraction), Ctrl. idx. (runtime control-activity index), and residual-active fraction are dimensionless.}
\label{tab:fight_smoke_appendix}
\footnotesize
\begin{adjustbox}{max width=\textwidth}
\begin{tabular}{lrrrrrr}
\toprule
Method & Ref./path RMS & Alt. RMS & Airspeed RMS & Ctrl. idx. & Viol. frac. & Residual active frac.\\
\midrule
Baseline & 177.140 & 100.880 & 8.703 & 13.532 & 0.003500 & 0.000\\
Q residual & 185.854 & 77.780 & 11.156 & 14.244 & 0.005000 & 0.078\\
HJB residual & 189.378 & 68.204 & 14.742 & 15.109 & 0.003333 & 0.194\\
\bottomrule
\end{tabular}
\end{adjustbox}
\end{table}


\end{document}